\documentclass[10pt,twocolumn,letterpaper]{article}
\usepackage[T1]{fontenc}
\usepackage[pagenumbers]{cvpr}
\usepackage{array}
\usepackage{microtype}
\definecolor{cvprblue}{rgb}{0.21,0.49,0.74}
\usepackage[pagebackref,breaklinks,colorlinks,allcolors=cvprblue]{hyperref}
\title{GateTrace: Where Do Crowd Crossings Go Missing Under Occlusion?}
\author{Ao He\qquad Qianchen Yu\qquad Vivian W. H. Wong\\University of Florida}
\hypersetup{pdftitle={GateTrace: Where Do Crowd Crossings Go Missing Under Occlusion?},
  pdfauthor={Ao He, Qianchen Yu, Vivian W. H. Wong}}
="[fonts/SourceCodePro-Regular.ttf]" at 9pt
\newlength{\GTusable}
\begin{document}
\maketitle
\raggedbottom
\begin{figure*}[t]
\centering
\includegraphics[width=\textwidth]{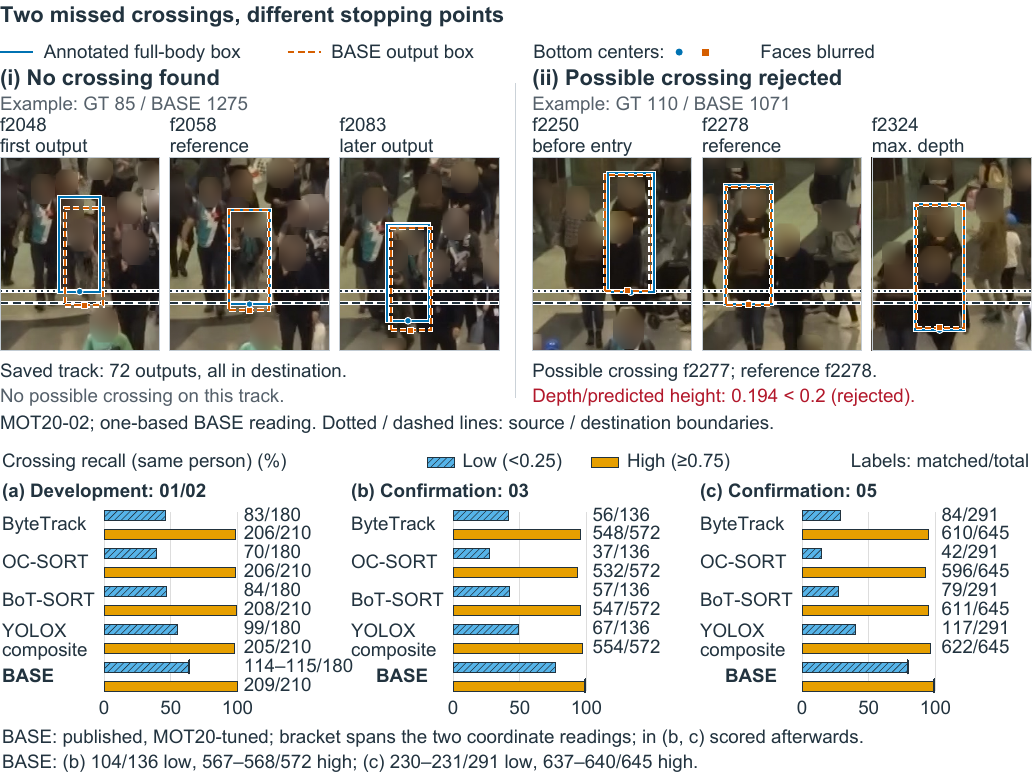}
\caption{Where crossings disappear from the reports. Top: two saved BASE examples from MOT20-02. (i) All 72 saved positions on the shown track lie in the destination region, so the counting rule finds no possible crossing. (ii) A possible crossing is found one frame before the reference, but the track does not enter the destination region far enough to pass the depth check: maximum depth divided by predicted body height is 0.194, below the 0.2 threshold. Faces are blurred. Bottom: crossing recall, the fraction of references matched by person, direction, time and position, in the lowest (\textless{}0.25) and highest (\ensuremath{\geq}0.75) visibility bins on the development sequences (a) and the two confirmation sequences (b, c). Labels give matched/total crossings. }
\label{fig:1}
\end{figure*}
\begin{abstract}
At large events, video-based tracking helps count people passing through entrances and passageways, but the final count does not show which crossings were missed or where they were lost. In a study of pedestrian flow through stadium tunnels, we found that the tracking system recorded about four in five crossings when pedestrians were clearly visible, but fewer than one in three when they were mostly obscured. The stadium videos lacked annotations that follow each person across frames, so we turned to MOT20, which provides annotated pedestrian trajectories and visibility. To locate these losses and assess changes to the counting rules, we introduce GateTrace. It derives reference crossings from annotated trajectories under fixed rules and matches reports by person, direction, time and position. For each miss, it checks for nearby boxes linked to the person, a possible crossing from those tracks, and rejection by later checks. With one shared detector, ByteTrack, OC-SORT and BoT-SORT recover 38.9--46.7\% of low-visibility reference crossings on the development sequences, compared with 98.1--99.0\% at high visibility. More than half of the low-visibility crossings therefore have no matching report. The gap recurs on two further sequences under a protocol fixed in advance. Published outputs of BASE, tuned on MOT20, still recover 63--78\% at low visibility across both sequence groups. The stages differ across systems. Across four systems and both sequence groups, 47--79\% of low-visibility misses have no nearby output linked to the person or no possible crossing in the saved tracks. Removing later checks from BASE on the development sequences leaves most misses unresolved and lowers the fraction of reports matching the fixed reference list from about 96\% to 77\%. Two human annotators checked sampled reference crossings against the videos; misses remain among the crossings they supported. GateTrace distinguishes these losses and measures the recovery and unmatched reports from changing later checks. We will release the code and per-crossing records.
\end{abstract}
\section{Introduction}

For tracking-based crowd-flow monitoring, a count begins with a person's trajectory through an entrance or passageway. A counting rule converts that trajectory into a report of when and in which direction the person crossed a virtual line. Occlusion can interrupt the trajectory or cause it to switch between people. A missing report then leaves no entry in the final count to inspect. To understand the omission, we need to identify the missing crossing and examine the steps that should have produced its report.

Our work was motivated by a video-based study of pedestrian flow at tunnel entrances in a football stadium. A tracker assigns each person a person-ID and tries to keep that ID across frames. In crowded scenes, however, we observed tracks switching between people or breaking into fragments, with crossings missing from the count. In comparisons by time and position, the YOLOX composite described in Sec.~4, adapted to the stadium's gate geometry and height model, matched 103 of 129 clearly visible crossings but only 40 of 138 crossings where the person was mostly obscured (Supplement F). The manual crossing annotations did not follow each person across frames, so we could not reliably check whether each report belonged to the correct person. To investigate these missed crossings, we turned to MOT20 \citep{ref2}, a public dataset of crowded scenes with the annotations needed for this check. These include pedestrian boxes, IDs that remain consistent across frames, and visibility values describing how much of each person can be seen.

GateTrace first lists the crossings derived from these annotations, then checks which ones have corresponding system reports. We define a region on either side of a virtual counting line; together, these form a counting gate. Fixed rules applied to annotated trajectories produce the reference list. On tracker outputs at the same gates, fixed rules find transitions between regions as possible crossings, which later checks retain or discard. A reference crossing counts as recovered when a system report matches it by person, direction, time and position. For every unmatched reference, GateTrace examines the saved boxes and counting decisions to locate where the report was lost.

Two examples from the BASE tracker in Fig.~\ref{fig:1} show why a missed crossing can require different checks. In the first, all saved positions on a track lie in the region the person entered. The counting procedure has no transition between regions to report. In the second, the track produces a possible crossing, but the check of how far it enters the destination region (the depth check) rejects it. Relaxing this check can affect the second case; the first still has no possible crossing to accept. GateTrace distinguishes these cases when evaluating changes to the counting rules.

We evaluate ByteTrack, OC-SORT and BoT-SORT \citep{ref3,ref4,ref5} with one shared YOLOX detector trained without MOT20 and the authors' MOT17 settings. On the development sequences, more than half of low-visibility reference crossings have no matching report, while almost all high-visibility crossings are recovered. The gap recurs on two further sequences under a protocol fixed in advance. Published outputs of BASE \citep{ref6}, whose parameters were fitted on MOT20, recover more low-visibility crossings but retain a gap between visibility groups. BASE's results on the two further sequences were scored after the fixed confirmation study.

These systems lose crossings at different stages. Across four systems and both sequence groups, the largest share of low-visibility misses has no nearby tracker output linked to the person or no possible crossing produced from that output. Later checks act on possible crossings already found in the saved tracks. These records therefore distinguish misses that later checks can affect from those that require an earlier change.

To test the effect of relaxing these checks, we keep BASE's tracks and the reference list fixed and remove all checks after possible crossings form. On the development sequences, this recovers only a small share of missed low-visibility crossings and adds unmatched reports. Precision, the fraction of all reports matching the fixed list, consequently falls.

These comparisons depend on the reference list. Full-body annotation boxes include occluded body parts, so a crossing derived from them can occur while the person's feet are hidden. Two annotators therefore reviewed the videos to judge whether each sampled crossing happened, in which direction, and whether the same person appeared on both sides. They assessed timing separately. Of the 18 low-visibility crossings both annotators initially supported on the first three questions, ByteTrack still misses eight and BASE three. Each evaluated system misses at most one of the 18 supported high-visibility crossings.

Our contributions are:

\begin{itemize}
\item GateTrace matches crossings from annotated tracks to reports and traces misses hidden by total counts.
\item Across four MOT20 sequences, three trackers sharing one detector miss over half of low-visibility crossings but few high-visibility ones. Published outputs of MOT20-tuned BASE still miss about 22--37\% of low-visibility crossings across both sequence groups. These results support evaluating crossing recovery by visibility.
\item Later checks offer limited recovery. In four systems on both sequence groups, 47--79\% of low-visibility misses lack nearby output linked to the person or a crossing found in saved tracks. Removing later checks on development data adds few low-visibility matches for BASE and ByteTrack and lowers precision against fixed references.
\end{itemize}

\section{Related work}

\textbf{Tracking evaluation in crowds.} MOTChallenge \citep{ref7} and MOT20 \citep{ref2} provide standard settings for pedestrian tracking, and HOTA \citep{ref8} balances detection, association and localization. MOTCOM \citep{ref11} describes sequence difficulty through occlusion, motion and appearance; CrowdTrack \citep{ref14} focuses on difficult pedestrian tracking in real scenes. Sundararaman et al. track pedestrian heads in dense crowds \citep{ref22}. Frame metrics grouped by visibility already show degradation under occlusion. We examine how this affects crossing reports and locate where each crossing is lost.

\textbf{Events and line crossing.} Roth et al. evaluated trackers on derived events and event recovery for each object \citep{ref9,ref23}. Ma and Chan counted line crossings with local features \citep{ref10}, AdaLine optimizes counting-line placement \citep{ref15}, and ByteTraX reports line counts \citep{ref16}. Pouw et al. benchmark pedestrian-flow and trajectory measurements \citep{ref1}. These studies establish crossing events as an evaluation target. Video individual counting estimates the number of distinct pedestrians from inflow and outflow between frames, without tracking \citep{ref24,ref25}. GateTrace evaluates tracker outputs by matching each crossing report to the annotated person, grouping results by visibility, and examining the tracking and counting records for every miss.

\textbf{Error analysis.} TIDE separates detection errors into categories \citep{ref17}, Detecting Invisible People evaluates interventions for occluded people \citep{ref18}, CAST examines association errors over time \citep{ref12}, and Perception Characteristics Distance relates perception quality to a decision rule \citep{ref13}. GateTrace applies this view to crowd-flow crossing reports.

\section{The GateTrace framework}

\begin{figure*}[t]

\centering

\includegraphics[width=\textwidth]{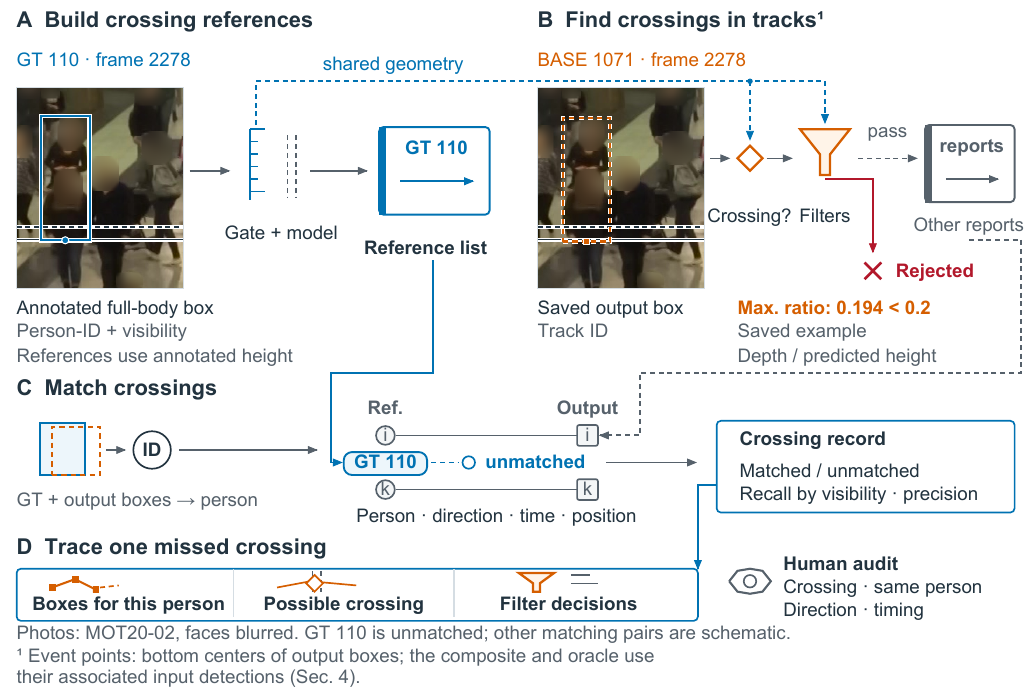}

\caption{How GateTrace checks crossing reports. (A) Annotated trajectories provide reference crossings and a height model for each sequence. (B) Counting rules find possible crossings in tracker outputs, check them and save the intermediate decisions. (C) Reports are matched to references by person, direction, time and position to calculate crossing recall and precision. (D) For each missed reference, a record connects nearby output boxes, possible crossings and counting decisions. Human judgments are reported separately.}
\label{fig:2}

\end{figure*}

GateTrace evaluates individual crossing reports in four steps (Fig.~\ref{fig:2}). It builds a reference list from annotated trajectories using fixed gate rules (Sec.~3.1), applies counting rules to each tracker's output at the same gates (Sec.~3.2), and matches reports to references by person, direction, time and position (Sec.~3.3). For each unmatched reference, it checks whether nearby boxes could be linked to that person, whether those tracks produced a possible crossing, and whether later checks kept it (Sec.~3.4).

\subsection{Crossing references}

We use MOT20 pedestrians with flag 1 and class 1. We represent each person's position by the bottom center of the annotated full-body box. This is the event point; points touching the left, right or lower image border are invalid. Each sequence has six gates: horizontal lines at $0.30H$, $0.50H$ and $0.70H$ and vertical lines at $0.25W$, $0.50W$ and $0.75W$, where $H$ and $W$ are image height and width. Gate sizes follow the apparent body height at their position. We fit an affine model of box height against image position for each sequence and predict the height $h_m$ at each gate midpoint. Each gate has two adjacent rectangular regions of depth $h_m$, separated by a gap of half-width $0.05h_m$.

We record a transition when a person first appears in the region opposite their last region assignment. A gap longer than 0.5 s resets that assignment; we do not interpolate missing positions. We include a transition in the reference list when a point reaches at least 0.2 times its annotated box height into the destination region between entry and 50 frames later, inclusive (Supplement F.3). We keep the excluded transitions for the analyses in Sec. 5.3. Crossing visibility is the person's mean annotated visibility within \ensuremath{\pm}25 frames (\ensuremath{\pm}1 s), grouped into $[0,.25)$, $[.25,.5)$, $[.5,.75)$ and $[.75,1]$. 

\subsection{Turning tracks into crossing reports}

We apply the counting rules to each output track, using the bottom centers of its boxes (except for the composite and oracle; Sec. 4). Each transition between the gate regions marks a possible crossing. We check and group these before reporting them.

The \emph{depth check} requires the track to reach 0.2 predicted body heights into the destination region within the next 2 s of the same track fragment. We merge same-track, same-direction events at most 2 s apart. For each group, we examine only valid points assigned to either region. In the one-second windows before and after the group, at least half must lie in the source and destination regions, respectively; groups with fewer than three in either window are kept and flagged. The remaining events are final crossing reports. We save these possible crossings, measured depths, groups and final decisions.

\subsection{Matching reports to the correct person}

To determine which annotated person a report describes, we use its track's boxes within \ensuremath{\pm}25 frames of the event. We match as many output and annotated boxes as possible one-to-one at IoU\ensuremath{\geq}0.5, where IoU is box intersection over union. We link a matched box to that person only when its overlap exceeds the best competing person's by at least 0.10. Each linked box votes for that person's ID. The event receives the most frequent person-ID if it has votes from at least half of all observations (at least three). If the IDs before and after the event conflict, we mark the person assignment as mixed.

A report can match a reference when person and direction agree, their times differ by at most 1 s, and their points are no farther apart than $h_g$, the person's annotated box height at the reference frame. For each sequence and gate, we maximize the number of matched pairs and then minimize their total time difference.

Recall measures the fraction of reference crossings recovered: in visibility bin $b$, it is $K_b/N_b$, where $N_b$ counts references and $K_b$ counts matched ones. Precision measures the fraction of final reports that match the reference list: it is $K/S$, where $K$ counts matched reports and $S$ counts all final reports, including those whose person assignment is unknown or mixed. Equally optimal assignments give identical bin counts when matching requires the same person (Supplement J.7).

\subsection{Tracing a missed crossing}

For each unmatched reference, GateTrace finds boxes linked to that person within \ensuremath{\pm}25 frames by the box-level check. From tracks containing these boxes, it retrieves possible crossings and checking records at that gate, including crossings outside this window.

GateTrace first searches for a final report for the same person and direction within the time and position limits. If none is found, it searches possible crossings that passed the depth check, then all possible crossings, under the same conditions. Those meeting these conditions but failing the depth check are recorded as rejected by that check.

If none meets these conditions, GateTrace checks the following cases in order and assigns the first that applies:
\begin{enumerate}
\item No nearby output box is linked to the person.
\item A possible crossing for the same person fails the direction, time or position conditions.
\item A possible crossing on a track containing boxes linked to the person meets those conditions, but its person assignment is not accepted for this reference.
\item A possible crossing fails to match for another reason.
\item Boxes linked to the person are present, but their tracks show no possible crossing at the gate.
\end{enumerate}

These records locate the step where a crossing was lost. Finding a possible crossing depends on the bottom-point rule, gate regions and gap reset. When a rule change adds reports, GateTrace records all additions, including unmatched reports and whether they correspond to annotated transitions outside the reference list.

\section{Experimental setup}

\textbf{Data split.} We develop the analysis on MOT20-01/02 (724 references; 180 low- and 210 high-visibility) and check its findings on MOT20-03/05 (3,948 references; 427 low- and 1,217 high-visibility). We call these the development and confirmation sets. After inspecting baseline results on 03/05, we fixed the confirmation protocol to constrain subsequent analysis choices.

\textbf{Systems.} We compare ByteTrack, OC-SORT and BoT-SORT \citep{ref3,ref4,ref5} using the same detections. We export detections once per sequence from the official ByteTrack YOLOX-X checkpoint \citep{ref3,ref19} (800\ensuremath{\times}1440 input, score 0.01, NMS 0.7). Its documented training data include CrowdHuman \citep{ref21} but exclude MOT20. We use the authors' MOT17 tracking settings; BoT-SORT runs without ReID unless stated.

The tuned \emph{YOLOX composite} admits detections at 0.10, runs ByteTrack at thresholds 0.25/0.10/0.25 and joins track fragments. BASE \citep{ref6} supplies published tracks produced with MOT20-based parameter fitting, access to later frames and interpolation. The released files do not document their coordinate origin, so we report both one- and zero-based readings.

The \emph{annotation-box oracle} supplies annotated boxes to the composite's tracker and counting rules. The composite and oracle take their event points from the input detections associated with each track (Supplement B.2). Person matching always uses output boxes. We keep the model weights fixed. CrowdDet-based systems \citep{ref20} appear in Supplements A and J.1; further variants appear in Supplement B.2.

\textbf{Fixed confirmation protocol.} Before running 03/05, we fixed two hypotheses for each sequence. H-A: each of these three trackers recovers at most 70\% of low-visibility and at least 90\% of high-visibility references. H-B: the composite recovers more low-visibility references than the best of these three trackers. Both must hold on both sequences. We later scored BASE's published outputs for 03/05 against the same references as an additional analysis. This falls outside the fixed confirmation; BASE's parameters were fitted on MOT20 training sequences that include these two.

\textbf{Human review.} Two annotators watched videos for 88 development references, sampled by visibility (34 low-visibility, 18 in each other bin). They judged whether the person crossed, in which direction, whether the person on both sides was the same, and when the crossing happened. A reference is supported when both say the crossing occurred, identify the same person on both sides, and agree with the reference direction. They first answered separately, then discussed disagreements; we report both sets of judgments. During initial answering, one annotator consulted an AI assistant about definitions. Their discussion interface displayed AI reference answers (Supplement C).

\section{Findings}

We measure missed crossings by visibility (Sec.~5.1), trace where they are lost (Sec.~5.2), and assess the recovery and precision cost of rule changes (Sec.~5.3). We then check human-supported crossings (Sec.~5.4) and compare frame metrics (Sec.~5.5).

\begin{table*}[t]
\centering
\caption{Development results (MOT20-01/02). Each visibility bin gives matched/total references. Precision is matched reports divided by all final reports. BASE values span the one- and zero-based coordinate readings.}
\label{tab:1}
\small
\setlength{\tabcolsep}{4pt}
\begin{tabular}{@{}lccccc@{}}
\toprule
System & Low & Mid-low & Mid-high & High & Precision \\
\midrule
ByteTrack & 83/180 & 142/171 & 149/163 & 206/210 & 580/644 \\
OC-SORT & 70/180 & 137/171 & 147/163 & 206/210 & 560/593 \\
BoT-SORT & 84/180 & 142/171 & 150/163 & 208/210 & 584/634 \\
BoT-SORT + ReID & 81/180 & 141/171 & 149/163 & 208/210 & 579/628 \\
YOLOX composite & 99/180 & 149/171 & 150/163 & 205/210 & 603/665 \\
BASE (published) & 114--115/180 & 155--156/171 & 151--152/163 & 209/210 & 629--632/659 \\
Annotation-box oracle & 173/180 & 168/171 & 161/163 & 209/210 & 711/723 \\
\bottomrule
\end{tabular}
\end{table*}

\subsection{Crossing recovery drops sharply at low visibility}

\textbf{Low-visibility crossings are often missed.} On the development set the three trackers sharing a detector recover 70--84/180 low-visibility and 206--208/210 high-visibility crossings (Table~\ref{tab:1}). The tuned composite reaches 99/180 and BASE 114--115/180, while high-visibility recovery stays at 205--209/210 for all systems. Precision stays between 90.1\% and 95.9\% for all non-oracle systems. The gap remains in four systems under all nine tested combinations of matching tolerances (0.5--1.5 s and 0.5--1.5 body heights; Supplement B.3).

The same pattern appears on the confirmation sequences (Table~\ref{tab:2}). H-A holds for all three trackers sharing a detector on both sequences, and H-B holds with 67 vs. 57 low-visibility matches on 03 and 117 vs. 84 on 05. Combining 03/05, low-visibility recovery is 18.5--32.8\% for these three trackers and 43.1\% for the composite, against 92.7--96.6\% for high-visibility crossings. In the additional analysis, BASE recovers 334--335/427 low-visibility and 1204--1208/1217 high-visibility crossings on 03/05, with a precision of 97.2\%. BASE's low-visibility recovery is 78.2--78.5\%, still about 21 points below high visibility.

\begin{table}[t]
\centering
\caption{Confirmation results (MOT20-03/05), evaluated under the protocol fixed before execution. BASE was scored afterwards as an additional analysis; its values span the two coordinate readings.}
\label{tab:2}
\fontsize{8.5}{10}\selectfont
\setlength{\tabcolsep}{2pt}
\begin{tabular}{@{}>{\raggedright\arraybackslash}p{0.24\columnwidth}cccc@{}}
\toprule
System & \shortstack{03\\Low} & \shortstack{03\\High} & \shortstack{05\\Low} & \shortstack{05\\High} \\
\midrule
ByteTrack & 56/136 & 548/572 & 84/291 & 610/645 \\
OC-SORT & 37/136 & 532/572 & 42/291 & 596/645 \\
BoT-SORT & 57/136 & 547/572 & 79/291 & 611/645 \\
YOLOX composite & 67/136 & 554/572 & 117/291 & 622/645 \\
BASE (published; additional) & 104/136 & 567--568/572 & 230--231/291 & 637--640/645 \\
Annotation-box oracle & 132/136 & 569/572 & 280/291 & 642/645 \\
\bottomrule
\end{tabular}
\end{table}

\subsection{Most BASE misses occur before later checks}

\textbf{Most low-visibility BASE misses occur before later checks.} Table~\ref{tab:3} groups the saved records for missed references by the step where they stop. Its first row accounts for BASE's 65--66 missed low-visibility crossings on the development sequences. For 20 crossings, no output box within \ensuremath{\pm}1 s can be linked to the person by the box-level check. Missed detections, inaccurate boxes or ambiguous person assignments could each produce this state. For 23, boxes are linked to the person, but their tracks show no possible crossing at the gate. For 13, a possible crossing meets the matching conditions but fails the depth check. Final reports and crossings passing that check failed the matching conditions for these misses. Thus none was lost through competition in matching, merging events or checking positions before and after the event.

\begin{table*}[t]
\centering
\caption{Where missed low-visibility crossings stop. The states defined in Sec. 3.4 are counted for four systems on the development (01/02) and confirmation (03/05) sequences. BASE entries give one-/zero-based readings where they differ. Nearby output means boxes linked to the person in the reference window. \emph{Other}: a possible crossing passed the depth check but was removed by merging events or checking positions before and after them, or failed to match for another reason on a track containing these boxes. No missed crossing has a compatible final report.}
\label{tab:3}
\small
\setlength{\tabcolsep}{4pt}
\begin{tabular}{@{}lcccccccc@{}}
\toprule
\shortstack{System} & \shortstack{Sequences} & \shortstack{No linked\\nearby\\output} & \shortstack{Linked output,\\no possible\\crossing} & \shortstack{Rejected by\\depth check} & \shortstack{Person\\assignment\\not accepted} & \shortstack{Direction, time\\or position\\mismatch} & \shortstack{Other} & \shortstack{Total\\missed} \\
\midrule
BASE & 01/02 & 20 & 23 & 13 & 5/4 & 5 & 0 & 66/65 \\
BASE & 03/05 & 26/25 & 28/29 & 25 & 8/10 & 5/4 & 0 & 92/93 \\
ByteTrack & 01/02 & 19 & 30 & 11 & 20 & 9 & 8 & 97 \\
OC-SORT & 01/02 & 24 & 42 & 16 & 9 & 10 & 9 & 110 \\
BoT-SORT & 01/02 & 18 & 27 & 12 & 21 & 9 & 9 & 96 \\
ByteTrack & 03/05 & 117 & 52 & 34 & 54 & 12 & 18 & 287 \\
OC-SORT & 03/05 & 165 & 111 & 37 & 17 & 8 & 10 & 348 \\
BoT-SORT & 03/05 & 118 & 64 & 38 & 48 & 13 & 10 & 291 \\
\bottomrule
\end{tabular}
\end{table*}

\textbf{The mix of losses varies across systems.} We examined the same records for the three trackers sharing a detector on both sequence groups and for BASE on 03/05 (Table~\ref{tab:3}). For 47--79\% of low-visibility misses, we find no nearby box linked to the person or no possible crossing. Together, these states account for more than half in every case except BoT-SORT on the development sequences (45 of 96). The depth check rejects 11--27\%. Another state is more common in ByteTrack and BoT-SORT: 16--22\% of their misses have a possible crossing that meets the time and position conditions but is not accepted as belonging to the person. This accounts for 6--11\% in BASE and 5--8\% in OC-SORT.

\begin{figure*}[t]

\centering

\includegraphics[width=\textwidth]{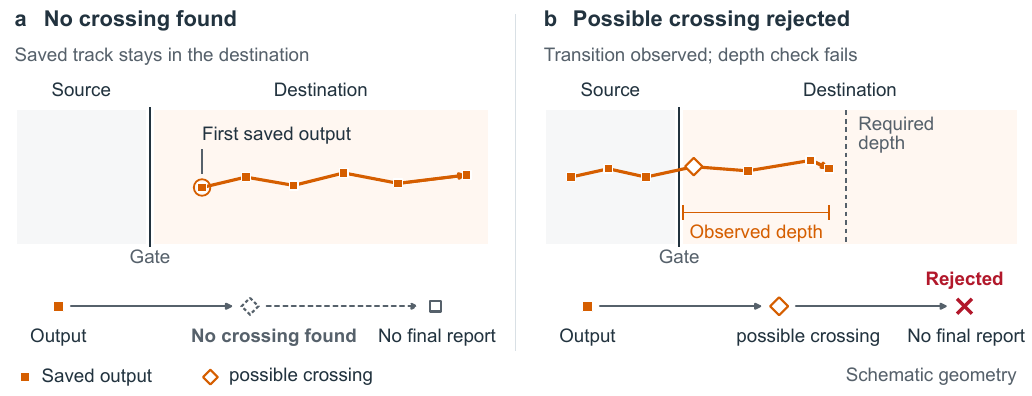}

\caption{Two steps at which a crossing can be lost. (a) The person's track stays entirely inside the destination region, so the counting rule finds no possible crossing. (b) A possible crossing is found but fails the depth check. Geometry is schematic; both cases come from the saved BASE records.}

\label{fig:3}

\end{figure*}

Figure~\ref{fig:3} traces the two BASE misses in Fig.~\ref{fig:1}. A continuing track's possible crossing fails the depth check: distance into the destination region divided by predicted body height is 0.194--0.199, below 0.2; using output box height gives 0.202--0.208. In the other case, 36 linked output boxes surround the reference, but the entire 72-frame track stays in the destination region without an internal gap. The counting rule finds no transition between regions for later checks to retain.

Two controls help interpret these misses (Supplement B.1). Supplying annotated boxes lets the composite recover nearly all low-visibility crossings. This changes observation availability and box geometry together and aligns crossing points with the references. Deleting half the annotated observations has a much larger effect when deletions are contiguous, reflecting both their timing and the reset of the last known region after a gap longer than 0.5 s.

\subsection{Changing later checks recovers few missed crossings}

\textbf{Filter changes add few low-visibility matches and lower precision.} We hold BASE's tracks and the original references fixed and remove the depth check, merging of events and checks of positions before and after them (Table~\ref{tab:4}). Matched low-visibility crossings rise from 114--115 to 127--128 of 180 and high-visibility recovery reaches 210/210; ByteTrack rises from 83 to 95. Against the same references, precision falls from 95.4--95.9\% to 76.6--77.0\% for BASE and from 90.1\% to 65.9\% for ByteTrack. The net gain of 13 counts changes in matches; the 13 depth-check rejections in Table~\ref{tab:3} count misses under the original rules.

\begin{table*}[t]
\centering
\caption{Changing the counting rules. Recall and precision use the same original development references in every row.}
\label{tab:4}
\small
\setlength{\tabcolsep}{4pt}
\begin{tabular}{@{}lcccc@{}}
\toprule
Output & Counting rules & Low & High & Precision \\
\midrule
BASE & Original & 114--115/180 & 209/210 & 95.4--95.9\% \\
BASE & Keep rejected possible crossings & 120/180 & 209/210 & 94.0--94.7\% \\
BASE & Relaxed (no filters) & 127--128/180 & 210/210 & 76.6--77.0\% \\
ByteTrack & Original & 83/180 & 206/210 & 90.1\% \\
ByteTrack & Relaxed (no filters) & 95/180 & 208/210 & 65.9\% \\
\bottomrule
\end{tabular}
\end{table*}

We also tested a selective change: keeping BASE's possible crossings rejected by the depth check when the YOLOX composite supports them. It adds five to six low-visibility matches and loses none, but also adds 9--11 unmatched reports, most of which coincide with annotated transitions outside the original reference list. We designed this exploratory repair after inspecting development failures. Its acceptance rule, recorded before the run, required no precision loss, so we do not adopt it. With these outputs and rules for finding possible crossings fixed, most low-visibility misses occur before the later filters: even without these filters, 52--53 of 180 remain unmatched for BASE and 85 for ByteTrack.

\textbf{The reference rule affects which reports match.} Some unmatched reports correspond to annotated transitions excluded from the original reference list. Keeping reports from the relaxed rules fixed and adding these transitions to the reference list gives 147--151 additional matches for BASE and 121 for ByteTrack, with precision 93.5--94.4\% and 78.9\% (Supplements G and J.5). These scores measure agreement with the expanded list of annotated transitions, whose physical occurrence requires separate evidence. Against the original list, the relaxed reports retain the lower precision reported above.

\subsection{Misses remain among crossings supported by human review}

\textbf{Human review supports crossings that systems still miss.} Both annotators initially supported 64 of 88 sampled references, including 18 of 34 low-visibility ones. After discussing disagreements, these counts were 79 and 25. Within the supported subsets, every system misses more low- than high-visibility crossings (Table~\ref{tab:5}); for ByteTrack the initial counts are 8 of 18 against 0 of 18.

\begin{table}[t]
\centering
\caption{Misses among crossings supported by both annotators. Each cell gives unmatched/supported references within the sample, without weighting.}
\label{tab:5}
\fontsize{8.5}{10}\selectfont
\setlength{\tabcolsep}{2pt}
\begin{tabular}{@{}>{\raggedright\arraybackslash}p{0.29\columnwidth}cccc@{}}
\toprule
System & \shortstack{Initial\\low} & \shortstack{Initial\\high} & \shortstack{Reconciled\\low} & \shortstack{Reconciled\\high} \\
\midrule
ByteTrack & 8/18 & 0/18 & 12/25 & 0/18 \\
OC-SORT & 10/18 & 1/18 & 16/25 & 1/18 \\
BoT-SORT & 9/18 & 0/18 & 12/25 & 0/18 \\
YOLOX composite & 3/18 & 1/18 & 5/25 & 1/18 \\
BASE & 3/18 & 0/18 & 6/25 & 0/18 \\
BASE + repair & 2/18 & 0/18 & 4/25 & 0/18 \\
\bottomrule
\end{tabular}
\end{table}

Timing is less certain. Of 14 sampled references for which both annotators gave a time interval, 13 intervals overlap. Only five pairs both contain the reference frame, and only one of those is low-visibility.

\subsection{Relation to frame-level metrics}

\textbf{Crossing recall measures omissions as missed passages.} Frame metrics also decline at low visibility. Only 3 of 56 confirmation-set comparisons give opposite tracker rankings by frame metrics and crossing recall, with differences meeting our thresholds: on MOT20-05, BoT-SORT scores higher on the frame metrics, but ByteTrack recovers five more low-visibility crossings (Supplement J.7).

\section{Limitations}

The study covers four MOT20 sequences from a small number of scene conditions, with many dependent crossings in each scene.  Three trackers share one detector; BASE uses published outputs with a different training and processing history and an undocumented coordinate origin. Reference points lie at the bottom of full-body annotation boxes and may be hidden behind occluders. Human judgments support the crossing, its direction and the same person on both sides in a sample drawn by visibility, but rarely exact timing. The saved records locate losses within the counting procedure. Several causes can produce the same recorded state: a late or broken track and differences between output and annotated box bottoms can each prevent the counting rule from finding a crossing. The tested repair and retrospective repair-selection check both failed their acceptance criteria (Supplement G).

\section{Conclusion}

GateTrace uses annotated trajectories, gate geometry, tracker outputs and saved counting records to match crossings to reports and trace omissions hidden by total counts. Across four MOT20 sequences, three trackers sharing one detector miss over half of low-visibility crossings; published outputs of MOT20-tuned BASE miss a fifth to over a third of these crossings across both sequence groups. Human video review also supports some crossings that remain unmatched. The mix of low-visibility losses differs across four systems, but the largest share lacks nearby output linked to the person or a possible crossing; fewer fail the depth check. Later checks process only crossings already found. Removing them on development data adds few low-visibility matches for BASE and ByteTrack and lowers precision against fixed references. For tracking-based crowd-flow monitoring, GateTrace supports offline evaluation that reports crossing recall by visibility and inspects individual misses alongside frame metrics.

\section*{Ethics statement}

The stadium recordings were obtained from a prior IRB-approved study and shared with the authors with permission for their use in this research. Only aggregate stadium results are reported, and the recordings are excluded from the released resource. The two annotators are authors and agreed to the use of their judgments in this study. The main evaluation uses MOT20 under CC BY-NC-SA 3.0, and derived labels retain the required attribution.

\label{end:main}
\section*{LLM Usage Statement}
Claude (Anthropic) and OpenAI Codex (called Astra in Supplement D) were used to retrieve and check literature, generate related-work material, generate and rewrite passages from author-provided text and ideas, propose interpretations and analysis plans, implement and run analyses and numerical checks, and produce figures through generated plotting code. Some assigned literature-search, coding and analysis tasks involved multiple steps completed by the tools before author review. The authors reviewed these outputs and checked the final analyses, figures, claims and text. Codex and a locally run Qwen3-VL-8B-Instruct generated visual answers for the auxiliary comparison in Supplement D. During initial answering, one of the two annotators consulted Codex on task interpretation and examples; joint reconciliation displayed Codex and Qwen answers for consultation. Initial and reconciled human records are reported separately. AI visual answers were not used as reference labels or inputs to event matching. Human-support and human--human agreement statistics were derived from the recorded human responses under these assistance conditions. Numerical recomputations checked the reported statistics using shared upstream records and components (Supplement I). The authors take responsibility for the reported work.
\clearpage
\emergencystretch=2em
\setcounter{dbltopnumber}{3}
\renewcommand{\dbltopfraction}{0.9}
\renewcommand{\textfraction}{0.08}
\twocolumn[{\centering\Large\bfseries GateTrace: Supplementary Material\par\vspace{1em}}]

\textbf{Terms used below.} BASE denotes the published outputs of the BASE tracker [6]. Counting rules examine tracks for possible crossings and decide which become reports. The depth check is C1; grouping events from the same track and direction is R1, and checking which regions the observations occupy before and after is R2. O/S denote the original and relaxed rules for deciding what counts as a crossing. Q2/Q3 are the human audits using samples selected by system outcomes and by sequence and visibility, respectively. P2/O1 denote the CrowdDet baseline and annotation-box oracle. The table in Appendix F.2 lists all project labels.

This supplement gives the complete result tables and the details condensed out of the main paper. Reference numbers refer to the main paper's bibliography. Appendix I describes which calculations were repeated from archived records and which inputs those checks shared. Appendix K reports two analyses run after the fixed confirmation: BASE on MOT20-03/05, and the stages at which further systems miss crossings.

\section*{Appendix A. Complete confirmation results}

\label{app:A}

\begin{table*}[!tp]
\centering
\small
\begin{minipage}{\textwidth}
Table A1 gives the matched/total counts for each sequence. All seven systems passed the output checks, and independent numerical recomputation agreed on both sequences. We judge the confirmation separately on each sequence; the combined rows describe their totals.
\end{minipage}\par\vspace{4pt}
\fontsize{9}{10.8}\selectfont
\setlength{\tabcolsep}{3pt}
\setlength{\GTusable}{\dimexpr\textwidth-12\tabcolsep\relax}
\renewcommand{\arraystretch}{1.12}
\begin{tabular}{@{}>{\raggedright\arraybackslash}p{0.090000\GTusable}>{\raggedright\arraybackslash}p{0.230000\GTusable}>{\raggedright\arraybackslash}p{0.120000\GTusable}>{\raggedright\arraybackslash}p{0.140000\GTusable}>{\raggedright\arraybackslash}p{0.140000\GTusable}>{\raggedright\arraybackslash}p{0.140000\GTusable}>{\raggedright\arraybackslash}p{0.140000\GTusable}@{}}
\toprule
\textbf{Sequence} & \textbf{System} & \textbf{Low} & \textbf{Mid-low} & \textbf{Mid-high} & \textbf{High} & \textbf{Precision} \\
\midrule
03 & YOLOX composite & 67/\allowbreak{}136 & 271/\allowbreak{}339 & 367/\allowbreak{}391 & 554/\allowbreak{}572 & 1259/\allowbreak{}1336 \\
03 & ByteTrack & 56/\allowbreak{}136 & 244/\allowbreak{}339 & 353/\allowbreak{}391 & 548/\allowbreak{}572 & 1201/\allowbreak{}1279 \\
03 & OC-SORT & 37/\allowbreak{}136 & 199/\allowbreak{}339 & 327/\allowbreak{}391 & 532/\allowbreak{}572 & 1095/\allowbreak{}1130 \\
03 & BoT-SORT & 57/\allowbreak{}136 & 246/\allowbreak{}339 & 356/\allowbreak{}391 & 547/\allowbreak{}572 & 1206/\allowbreak{}1270 \\
03 & P2 & 43/\allowbreak{}136 & 169/\allowbreak{}339 & 301/\allowbreak{}391 & 474/\allowbreak{}572 & 987/\allowbreak{}1141 \\
03 & CrowdDet composite & 43/\allowbreak{}136 & 165/\allowbreak{}339 & 302/\allowbreak{}391 & 480/\allowbreak{}572 & 990/\allowbreak{}1169 \\
03 & O1 & 132/\allowbreak{}136 & 336/\allowbreak{}339 & 389/\allowbreak{}391 & 569/\allowbreak{}572 & 1426/\allowbreak{}1440 \\
05 & YOLOX composite & 117/\allowbreak{}291 & 667/\allowbreak{}804 & 712/\allowbreak{}770 & 622/\allowbreak{}645 & 2118/\allowbreak{}2275 \\
05 & ByteTrack & 84/\allowbreak{}291 & 574/\allowbreak{}804 & 667/\allowbreak{}770 & 610/\allowbreak{}645 & 1935/\allowbreak{}2114 \\
05 & OC-SORT & 42/\allowbreak{}291 & 437/\allowbreak{}804 & 594/\allowbreak{}770 & 596/\allowbreak{}645 & 1669/\allowbreak{}1729 \\
05 & BoT-SORT & 79/\allowbreak{}291 & 581/\allowbreak{}804 & 678/\allowbreak{}770 & 611/\allowbreak{}645 & 1949/\allowbreak{}2086 \\
05 & P2 & 39/\allowbreak{}291 & 339/\allowbreak{}804 & 522/\allowbreak{}770 & 528/\allowbreak{}645 & 1428/\allowbreak{}1962 \\
05 & CrowdDet composite & 36/\allowbreak{}291 & 335/\allowbreak{}804 & 520/\allowbreak{}770 & 521/\allowbreak{}645 & 1412/\allowbreak{}2051 \\
05 & O1 & 280/\allowbreak{}291 & 801/\allowbreak{}804 & 760/\allowbreak{}770 & 642/\allowbreak{}645 & 2483/\allowbreak{}2508 \\
03+05 & YOLOX composite & 184/\allowbreak{}427 & 938/\allowbreak{}1143 & 1079/\allowbreak{}1161 & 1176/\allowbreak{}1217 & 3377/\allowbreak{}3611 \\
03+05 & ByteTrack & 140/\allowbreak{}427 & 818/\allowbreak{}1143 & 1020/\allowbreak{}1161 & 1158/\allowbreak{}1217 & 3136/\allowbreak{}3393 \\
03+05 & OC-SORT & 79/\allowbreak{}427 & 636/\allowbreak{}1143 & 921/\allowbreak{}1161 & 1128/\allowbreak{}1217 & 2764/\allowbreak{}2859 \\
03+05 & BoT-SORT & 136/\allowbreak{}427 & 827/\allowbreak{}1143 & 1034/\allowbreak{}1161 & 1158/\allowbreak{}1217 & 3155/\allowbreak{}3356 \\
03+05 & P2 & 82/\allowbreak{}427 & 508/\allowbreak{}1143 & 823/\allowbreak{}1161 & 1002/\allowbreak{}1217 & 2415/\allowbreak{}3103 \\
03+05 & CrowdDet composite & 79/\allowbreak{}427 & 500/\allowbreak{}1143 & 822/\allowbreak{}1161 & 1001/\allowbreak{}1217 & 2402/\allowbreak{}3220 \\
03+05 & O1 & 412/\allowbreak{}427 & 1137/\allowbreak{}1143 & 1149/\allowbreak{}1161 & 1211/\allowbreak{}1217 & 3909/\allowbreak{}3948 \\
\bottomrule
\end{tabular}
\end{table*}

For H-A, the three low-visibility matched counts on 03 are 56,37,57 out of N=136; their high-visibility counts are 548,532,547 out of N=572. On 05 the corresponding counts are 84,42,79 out of N=291 and 610,596,611 out of N=645. Each satisfies the inequalities in main-paper Section 4; Appendix J.2 gives their exact integer form. H-B holds because 67\textgreater{}57 and 117\textgreater{}84. The CrowdDet comparison is also retained: P2 and the composite recover 43 versus 43 low-visibility events on 03, and 39 versus 36 on 05. Combining M3/M4 therefore gave no low-visibility gain in this confirmation comparison.

\section*{Appendix B. Controls, sensitivity, and uncertainty}

\label{app:B}

\subsection*{B.1. Attribution and controlled deletion}

Two controls help interpret these misses. When given annotated boxes, the composite's tracker and the same counting rules recover 173/180 low-visibility development crossings and 412/427 on the confirmation set. This oracle changes both the available observations and box geometry, and its crossing points coincide with the references. It shows that the counting rules can recover nearly all low-visibility crossings from this input, while leaving the contributions of these factors unresolved. A second control deletes half of each person's annotated observations. Overall crossing recall is 89.6\% when deletions are dispersed and 22.4--22.5\% when they are contiguous. This difference reflects both the timing of the missing observations and the rule that resets region history after a gap longer than 0.5 s.

\begin{table*}[!tp]
\centering
\small
\begin{minipage}{\textwidth}
E1 assigns each case in the saved outputs to the first applicable category in a fixed order. Table B1 counts each case in one category; U collects unresolved cases. When observations before and after an event identify different people, the rule overrides an otherwise assigned event person-ID. This affects 29 records with unresolved person-IDs across P1/P2/CrowdDet composite: 5/9/15. The table follows this documented rule, and independent recomputation agreed.
\end{minipage}\par\vspace{4pt}
\fontsize{9}{10.8}\selectfont
\setlength{\tabcolsep}{3pt}
\setlength{\GTusable}{\dimexpr\textwidth-16\tabcolsep\relax}
\renewcommand{\arraystretch}{1.12}
\begin{tabular}{@{}>{\raggedright\arraybackslash}p{0.180000\GTusable}>{\raggedright\arraybackslash}p{0.060000\GTusable}>{\raggedright\arraybackslash}p{0.060000\GTusable}>{\raggedright\arraybackslash}p{0.090000\GTusable}>{\raggedright\arraybackslash}p{0.150000\GTusable}>{\raggedright\arraybackslash}p{0.120000\GTusable}>{\raggedright\arraybackslash}p{0.120000\GTusable}>{\raggedright\arraybackslash}p{0.140000\GTusable}>{\raggedright\arraybackslash}p{0.080000\GTusable}@{}}
\toprule
\textbf{System} & \textbf{Bin} & \textbf{N} & \textbf{Recovered} & \textbf{No compatible input after admission} & \textbf{No supporting tracker output} & \textbf{Insufficient connections between observations} & \textbf{Counting rules /\allowbreak{} position} & \textbf{U} \\
\midrule
P2 & All & 724 & 566 & 4 & 1 & 30 & 31 & 92 \\
P2 & Low & 180 & 81 & 4 & 1 & 22 & 13 & 59 \\
CrowdDet composite & All & 724 & 569 & 4 & 3 & 28 & 30 & 90 \\
CrowdDet composite & Low & 180 & 86 & 4 & 3 & 23 & 12 & 52 \\
O1 & All & 724 & 711 & 0 & 0 & 0 & 7 & 6 \\
O1 & Low & 180 & 173 & 0 & 0 & 0 & 2 & 5 \\
\bottomrule
\end{tabular}
\end{table*}

\begin{table*}[!tp]
\centering
\small
\begin{minipage}{\textwidth}
E2 deletes an equal fixed number of observations from each continuous segment defined by person and visibility. Dispersed deletion samples positions without replacement; block deletion removes consecutive observations. Table B2 gives means over three deletion seeds at requested retention 0.5. The independent check covered 48 masks, 50 runs, and 250 scoring cells. It reused the tracker and counting rules, while separately implementing the deletion masks, input adapter and scoring. Before the full comparison, its protocol had shown two approximate summary values for retention 0.5.
\end{minipage}\par\vspace{4pt}
\fontsize{9}{10.8}\selectfont
\setlength{\tabcolsep}{3pt}
\setlength{\GTusable}{\dimexpr\textwidth-4\tabcolsep\relax}
\renewcommand{\arraystretch}{1.12}
\begin{tabular}{@{}>{\raggedright\arraybackslash}p{0.180000\GTusable}>{\raggedright\arraybackslash}p{0.410000\GTusable}>{\raggedright\arraybackslash}p{0.410000\GTusable}@{}}
\toprule
\textbf{Sequence} & \textbf{Dispersed deletion,\allowbreak{} overall recall} & \textbf{Contiguous deletion,\allowbreak{} overall recall} \\
\midrule
01 & 0.896396 & 0.225225 \\
02 & 0.896410 & 0.224103 \\
\bottomrule
\end{tabular}
\end{table*}

Rounding within each segment makes the retained fraction differ from the requested value. Both deletion patterns remove the same number of observations from each segment. No predictive formula was fixed in advance for these descriptive curves. Their values reflect both tracking behavior and the reset in the counting rules after a gap longer than 0.5 s.

\subsection*{B.2. Tested interventions and component comparisons}

\begin{table*}[!tp]
\centering
\small
\fontsize{9}{10.8}\selectfont
\setlength{\tabcolsep}{3pt}
\setlength{\GTusable}{\dimexpr\textwidth-4\tabcolsep\relax}
\renewcommand{\arraystretch}{1.12}
\begin{tabular}{@{}>{\raggedright\arraybackslash}p{0.230000\GTusable}>{\raggedright\arraybackslash}p{0.320000\GTusable}>{\raggedright\arraybackslash}p{0.450000\GTusable}@{}}
\toprule
\textbf{Analysis} & \textbf{Recorded outcome} & \textbf{Interpretation and verification} \\
\midrule
HeadGate,\allowbreak{} seven fixed diagnostic conditions & Low counts 20,\allowbreak{}20,\allowbreak{}21,\allowbreak{}21,\allowbreak{}21,\allowbreak{}20,\allowbreak{}20 out of 32; all high 12/\allowbreak{}13; required low 26/\allowbreak{}32 & Idealized head observations,\allowbreak{} person-IDs and positions under the fixed projection and counting rules; independent recomputation agreed \\
Screening possible crossings across gaps & P2 10; CrowdDet composite 16; required gain 17 & This test stopped after counting possible crossings; those sets were independently rebuilt \\
Relinking with reference person-IDs,\allowbreak{} two-second limit & ByteTrack +6; OC-SORT +21; BoT-SORT +9 low-bin matches & Required at least two systems with +27; independent transformation checked with the same scorer; test stopped \\
Fixed BoT-SORT ReID & Low 84\ensuremath{\to}81; high 208\ensuremath{\to}208; precision 579/\allowbreak{}628 & Independently recomputed; retained as a baseline comparison \\
YOLOX composite without M4 & Low 97/\allowbreak{}180; high 205/\allowbreak{}210; precision 600/\allowbreak{}657 & Independent recomputation agreed \\
Composite with points from output boxes & Low 98/\allowbreak{}180; high 205/\allowbreak{}210; precision 602/\allowbreak{}664 & Saved-event scores reproduced by an additional implementation; events and assignments linking boxes to people in each frame shared \\
Author high/\allowbreak{}new thresholds,\allowbreak{} without M3/\allowbreak{}M4 & Low 86/\allowbreak{}180; high 206/\allowbreak{}210; precision 587/\allowbreak{}642 & Independent recomputation agreed \\
\bottomrule
\end{tabular}
\end{table*}

HeadGate uses head annotations from the same MOT20-01 video; Sundararaman et al. [22] describe their dense-head benchmark source. The relinking test keeps the saved observations and the specified gap and interpolation procedure fixed. The component comparisons change several parts of the processing chain. The full composite gains 13 events over the author-threshold variant and two over the variant without M4. These differences describe the tested configurations; they cannot allocate all missed crossings additively to individual components.

\subsection*{B.3. Existing descriptive robustness analyses}

The saved tolerance checks cover four systems at nine time/distance combinations. Nominal time thresholds of 0.5,1.0,1.5 s correspond to 12,25,37 frame ticks at 25 fps; spatial thresholds are 0.5,1.0,1.5 reference heights. The low/high visibility gap remains across the reported 144 rows. Increasing the event-matching time tolerance from 1.0 to 1.5 s adds at most three low-visibility matches. We also checked visibility defined by the minimum over \ensuremath{\pm}25 frames, the mean over \ensuremath{\pm}12, or the value at the reference frame. The gap remains when the pooled samples are divided into density thirds. The source analysis records ties and undefined values in very small groups.

\begin{table*}[!tp]
\centering
\small
\begin{minipage}{\textwidth}
The person-cluster bootstrap resamples people within each sequence, using 2,000 paired resamples, seed 20261025, and 95\% percentile intervals. These descriptive results were computed after the main analysis and have not been checked with a second, independent bootstrap implementation. The scenes, references, gates, and cached outputs are held fixed.
\end{minipage}\par\vspace{4pt}
\fontsize{9}{10.8}\selectfont
\setlength{\tabcolsep}{3pt}
\setlength{\GTusable}{\dimexpr\textwidth-4\tabcolsep\relax}
\renewcommand{\arraystretch}{1.12}
\begin{tabular}{@{}>{\raggedright\arraybackslash}p{0.480000\GTusable}>{\raggedright\arraybackslash}p{0.270000\GTusable}>{\raggedright\arraybackslash}p{0.250000\GTusable}@{}}
\toprule
\textbf{Low-bin comparison} & \textbf{Difference in percentage points} & \textbf{95\% descriptive interval} \\
\midrule
03: YOLOX composite \ensuremath{-} BoT-SORT & 7.35 & [1.92,\allowbreak{}12.98] \\
05: YOLOX composite \ensuremath{-} ByteTrack & 11.34 & [6.91,\allowbreak{}15.92] \\
Pooled: composite \ensuremath{-} best of the three trackers sharing detections on each sequence & 10.07 & [6.59,\allowbreak{}13.76] \\
Pooled: composite \ensuremath{-} ByteTrack & 10.30 & [6.78,\allowbreak{}14.11] \\
\bottomrule
\end{tabular}
\end{table*}

Selecting the best of the three trackers sharing detections separately on each sequence gives 57+84=141 low-visibility matches in total. ByteTrack gives 56+84=140. These are separate comparisons.

\section*{Appendix C. Human judgments and reference support}

\label{app:C}

\begin{table*}[!tp]
\centering
\small
\begin{minipage}{\textwidth}
\textbf{Q3 sampling fractions by sequence and visibility.} N counts eligible references retained by the reference depth rule after excluding Q2; n counts sampled references from that list. The eight fractions describe the final sampling design. The human results below are unweighted counts from the observed subsets.
\end{minipage}\par\vspace{4pt}
\fontsize{9}{10.8}\selectfont
\setlength{\tabcolsep}{3pt}
\setlength{\GTusable}{\dimexpr\textwidth-8\tabcolsep\relax}
\renewcommand{\arraystretch}{1.12}
\begin{tabular}{@{}>{\raggedright\arraybackslash}p{0.250000\GTusable}>{\raggedright\arraybackslash}p{0.220000\GTusable}>{\raggedright\arraybackslash}p{0.190000\GTusable}>{\raggedright\arraybackslash}p{0.180000\GTusable}>{\raggedright\arraybackslash}p{0.160000\GTusable}@{}}
\toprule
\textbf{Sequence} & \textbf{Visibility} & \textbf{Eligible N} & \textbf{Sampled n} & \textbf{n/\allowbreak{}N} \\
\midrule
MOT20-01 & Low & 17 & 9 & 9/\allowbreak{}17 \\
MOT20-01 & Mid-low & 16 & 9 & 9/\allowbreak{}16 \\
MOT20-01 & Mid-high & 13 & 9 & 9/\allowbreak{}13 \\
MOT20-01 & High & 13 & 9 & 9/\allowbreak{}13 \\
MOT20-02 & Low & 83 & 25 & 25/\allowbreak{}83 \\
MOT20-02 & Mid-low & 155 & 9 & 9/\allowbreak{}155 \\
MOT20-02 & Mid-high & 150 & 9 & 9/\allowbreak{}150 \\
MOT20-02 & High & 197 & 9 & 9/\allowbreak{}197 \\
\bottomrule
\end{tabular}
\end{table*}

\begin{table*}[!tp]
\centering
\small
\begin{minipage}{\textwidth}
\textbf{Human support for sampled references retained by the depth rule.} Each triple gives supported / conflict / unresolved counts. Support requires agreement that the crossing occurred, that the person remained the same, and that the direction matches the reference. Initial answers and answers from joint review are reported separately.
\end{minipage}\par\vspace{4pt}
\fontsize{9}{10.8}\selectfont
\setlength{\tabcolsep}{3pt}
\setlength{\GTusable}{\dimexpr\textwidth-6\tabcolsep\relax}
\renewcommand{\arraystretch}{1.12}
\begin{tabular}{@{}>{\raggedright\arraybackslash}p{0.230000\GTusable}>{\raggedright\arraybackslash}p{0.150000\GTusable}>{\raggedright\arraybackslash}p{0.310000\GTusable}>{\raggedright\arraybackslash}p{0.310000\GTusable}@{}}
\toprule
\textbf{Visibility} & \textbf{Sample} & \textbf{Initial records} & \textbf{Joint review} \\
\midrule
Low & 34 & 18 /\allowbreak{} 8 /\allowbreak{} 8 & 25 /\allowbreak{} 5 /\allowbreak{} 4 \\
Mid-low & 18 & 12 /\allowbreak{} 2 /\allowbreak{} 4 & 18 /\allowbreak{} 0 /\allowbreak{} 0 \\
Mid-high & 18 & 16 /\allowbreak{} 1 /\allowbreak{} 1 & 18 /\allowbreak{} 0 /\allowbreak{} 0 \\
High & 18 & 18 /\allowbreak{} 0 /\allowbreak{} 0 & 18 /\allowbreak{} 0 /\allowbreak{} 0 \\
All & 88 & 64 /\allowbreak{} 11 /\allowbreak{} 13 & 79 /\allowbreak{} 5 /\allowbreak{} 4 \\
\bottomrule
\end{tabular}
\end{table*}

The initially supported low-visibility subset contains three events from MOT20-01 and fifteen from MOT20-02, involving eighteen distinct people. After joint review, these counts are five and twenty, involving twenty-five distinct people. The eighteen supported high-visibility events involve sixteen people. Events can still depend on one another through shared scenes and processing. These counts describe the sampled references; unequal sampling fractions, Q2 exclusion, and unresolved cases limit conclusions about the full population.

\begin{table*}[!tp]
\centering
\small
\begin{minipage}{\textwidth}
\textbf{Coverage of recorded Q3 timing intervals.} Counts use the 88 sampled references retained by the depth rule. \textquotedblleft{}Contains reference\textquotedblright{} requires both people's intervals to contain the reference frame. \textquotedblleft{}Excludes reference\textquotedblright{} means at least one interval excludes it. These interval counts did not change after joint review.
\end{minipage}\par\vspace{4pt}
\fontsize{9}{10.8}\selectfont
\setlength{\tabcolsep}{3pt}
\setlength{\GTusable}{\dimexpr\textwidth-14\tabcolsep\relax}
\renewcommand{\arraystretch}{1.12}
\begin{tabular}{@{}>{\raggedright\arraybackslash}p{0.130000\GTusable}>{\raggedright\arraybackslash}p{0.090000\GTusable}>{\raggedright\arraybackslash}p{0.110000\GTusable}>{\raggedright\arraybackslash}p{0.110000\GTusable}>{\raggedright\arraybackslash}p{0.140000\GTusable}>{\raggedright\arraybackslash}p{0.140000\GTusable}>{\raggedright\arraybackslash}p{0.140000\GTusable}>{\raggedright\arraybackslash}p{0.140000\GTusable}@{}}
\toprule
\textbf{Visibility} & \textbf{Sample} & \textbf{A interval} & \textbf{B interval} & \textbf{Both intervals} & \textbf{Intervals overlap} & \textbf{Contains reference} & \textbf{Excludes reference} \\
\midrule
Low & 34 & 5 & 25 & 4 & 4 & 1 & 4 \\
All & 88 & 16 & 75 & 14 & 13 & 5 & 12 \\
\bottomrule
\end{tabular}
\end{table*}

BASE recovered the single low-visibility item for which both intervals contain the reference frame. The intervals span 1.36 and 3.40 seconds, so they do not validate timing at the \ensuremath{\pm}1-second matching tolerance. Excluding supported items whose intervals explicitly conflict with the reference leaves initial low-visibility misses of 2/16 for BASE and 1/16 for the repair. These subsets still contain items with missing intervals. The human judgments support the reported subsets for occurrence, person and direction, with reference times and matching tolerances held fixed. Reference accuracy across the full population and BASE's miss rate on crossings with validated timing remain unresolved.

We recomputed the counts in this appendix from the original and formally revised answer files, saved sample mappings, and full-sequence matching pairs. The display-log account below identifies the export version used. The two people who reviewed the videos, A and B, submitted their initial answers separately before joint reconciliation. During initial completion, person A discussed task interpretation and illustrated examples with OpenAI Codex. We retain both the original judgments and the jointly reviewed answers. The authors administratively confirmed one item missed at submission while preserving its original in-progress state.

\begin{table*}[!tp]
\centering
\small
\begin{minipage}{\textwidth}
\textbf{Table C1. Agreement in the initial separately submitted records and after joint review.} Each entry counts identical answers, including identical uncertain answers. Q3 has 92 target items; its open-search item is excluded from these four target questions. Pooled rows count agreeing answers across the stated questions. Several answers come from each item, so these question-level counts are dependent. During initial completion, person A received the task-interpretation and example guidance described above and in main-paper Section 4. Joint review involved discussion with AI answers available for reference.
\end{minipage}\par\vspace{4pt}
\fontsize{9}{10.8}\selectfont
\setlength{\tabcolsep}{3pt}
\setlength{\GTusable}{\dimexpr\textwidth-4\tabcolsep\relax}
\renewcommand{\arraystretch}{1.12}
\begin{tabular}{@{}>{\raggedright\arraybackslash}p{0.460000\GTusable}>{\raggedright\arraybackslash}p{0.270000\GTusable}>{\raggedright\arraybackslash}p{0.270000\GTusable}@{}}
\toprule
\textbf{Question} & \textbf{Initial separately submitted} & \textbf{After joint review} \\
\midrule
Q2: body part visible before & 66/\allowbreak{}80 & 77/\allowbreak{}80 \\
Q2: body part visible after & 70/\allowbreak{}80 & 78/\allowbreak{}80 \\
Q2: crossing & 56/\allowbreak{}80 & 75/\allowbreak{}80 \\
Q2: pooled agreement across three questions & 192/\allowbreak{}240 (80.0\%) & 230/\allowbreak{}240 (95.83\%) \\
Q3: occurrence & 76/\allowbreak{}92 & 89/\allowbreak{}92 \\
Q3: direction & 72/\allowbreak{}92 & 90/\allowbreak{}92 \\
Q3: same person & 82/\allowbreak{}92 & 91/\allowbreak{}92 \\
Q3: timing determinable & 80/\allowbreak{}92 & 91/\allowbreak{}92 \\
Q3: pooled agreement across four questions & 310/\allowbreak{}368 (84.24\%) & 361/\allowbreak{}368 (98.10\%) \\
\bottomrule
\end{tabular}
\end{table*}

The pooled counts measure agreement across repeated questions about the same items. They differ from the number of items on which every answer agrees, and agreement alone does not establish that a reference is correct. Joint discussion increased agreement, while the supporting frame intervals stayed unchanged. Answers from OpenAI Codex and Qwen3-VL-8B-Instruct were available in an expandable display during discussion. The original follow-up export records 59 expansions of that display on 55 items: 34 items had an expansion before the first complete saved pair of follow-up answers, 18 after, and three still lacked a complete saved pair. Documented corrections subsequently completed those three. Counting their administrative completion timestamps gives 37 items; the underlying display events are unchanged. We retain the display logs and the authors' confirmation that they completed the final follow-up together.

\begin{table*}[!tp]
\centering
\small
\begin{minipage}{\textwidth}
\textbf{Table C2. Support for references retained by the depth rule.} A conflict is an explicit disagreement with the crossing's occurrence, the person involved, or the reference direction. Unresolved cases have uncertain or incomplete support.
\end{minipage}\par\vspace{4pt}
\fontsize{9}{10.8}\selectfont
\setlength{\tabcolsep}{3pt}
\setlength{\GTusable}{\dimexpr\textwidth-10\tabcolsep\relax}
\renewcommand{\arraystretch}{1.12}
\begin{tabular}{@{}>{\raggedright\arraybackslash}p{0.220000\GTusable}>{\raggedright\arraybackslash}p{0.140000\GTusable}>{\raggedright\arraybackslash}p{0.180000\GTusable}>{\raggedright\arraybackslash}p{0.150000\GTusable}>{\raggedright\arraybackslash}p{0.150000\GTusable}>{\raggedright\arraybackslash}p{0.160000\GTusable}@{}}
\toprule
\textbf{View} & \textbf{Bin} & \textbf{Sampled references} & \textbf{Supported} & \textbf{Conflict} & \textbf{Unresolved} \\
\midrule
Initial & Low & 34 & 18 & 8 & 8 \\
Initial & Mid-low & 18 & 12 & 2 & 4 \\
Initial & Mid-high & 18 & 16 & 1 & 1 \\
Initial & High & 18 & 18 & 0 & 0 \\
Initial & All & 88 & 64 & 11 & 13 \\
Joint review & Low & 34 & 25 & 5 & 4 \\
Joint review & Mid-low & 18 & 18 & 0 & 0 \\
Joint review & Mid-high & 18 & 18 & 0 & 0 \\
Joint review & High & 18 & 18 & 0 & 0 \\
Joint review & All & 88 & 79 & 5 & 4 \\
\bottomrule
\end{tabular}
\end{table*}

Q3's sample from the reference list excludes Q2, and sampling fractions differ by sequence and visibility. The table describes this sample without giving an error bound for the full population. For low-visibility items, A supplied timing intervals for five and B for 25. Four items have intervals from both; all four pairs overlap, and one pair contains the reference frame in both intervals. Across all bins the corresponding counts are 16,75,14,13,5. Twelve sampled references have at least one interval excluding the reference frame, including four low-visibility items.

\begin{table*}[!tp]
\centering
\small
\begin{minipage}{\textwidth}
\textbf{Table C3. Q2 support and BASE recovery under the two coordinate interpretations.} The group missed by all five earlier systems was defined before BASE entered the study. Ranges show how counts differ between coordinate interpretations.
\end{minipage}\par\vspace{4pt}
\fontsize{9}{10.8}\selectfont
\setlength{\tabcolsep}{3pt}
\setlength{\GTusable}{\dimexpr\textwidth-10\tabcolsep\relax}
\renewcommand{\arraystretch}{1.12}
\begin{tabular}{@{}>{\raggedright\arraybackslash}p{0.160000\GTusable}>{\raggedright\arraybackslash}p{0.250000\GTusable}>{\raggedright\arraybackslash}p{0.090000\GTusable}>{\raggedright\arraybackslash}p{0.210000\GTusable}>{\raggedright\arraybackslash}p{0.145000\GTusable}>{\raggedright\arraybackslash}p{0.145000\GTusable}@{}}
\toprule
\textbf{Human view} & \textbf{Original sample group} & \textbf{Sample} & \textbf{Both support crossing} & \textbf{BASE misses} & \textbf{Repair misses} \\
\midrule
Initial & Five-system common miss & 40 & 18 & 8--9/\allowbreak{}18 & 8--9/\allowbreak{}18 \\
Initial & At least three recovered & 40 & 35 & 3/\allowbreak{}35 & 2/\allowbreak{}35 \\
Joint review & Five-system common miss & 40 & 27 & 13--14/\allowbreak{}27 & 13--14/\allowbreak{}27 \\
Joint review & At least three recovered & 40 & 40 & 3--4/\allowbreak{}40 & 2/\allowbreak{}40 \\
\bottomrule
\end{tabular}
\end{table*}

The remaining items missed by all five systems keep their unsupported or uncertain crossing judgments. Both people reported complete invisibility for zero supported items in each row. Q2's original dominance criterion requires at least twenty supported crossings per group. With Y definite invisible-side judgments and U uncertain judgments among n supported crossings, the bounds are L=Y/n and U\_b=(Y+U)/n. The criterion also requires L\_common\textgreater{}0.50 and L\_common\ensuremath{-}U\_b,control\ensuremath{\geq}0.20. All three conditions must hold. The recorded interpretation failed this criterion.

\section*{Appendix D. Auxiliary human--AI comparison}

\label{app:D}

\begin{table*}[!tp]
\centering
\small
\begin{minipage}{\textwidth}
These counts report pairwise agreement among the two people who reviewed the videos and the two AI assessors, using the initial human answers and AI-generated visual answers. Astra denotes OpenAI Codex. It reviewed fixed sampled views, target cues, and some sequences of nearby frames, having already seen project context and screenshots. Qwen3-VL-8B-Instruct, run locally, completed 80 Q2 items and 82 Q3 items. Eleven Q3 items were missing because of a CUDA memory failure. The people who reviewed the videos could watch complete clips, pause, step through frames, and zoom. The assessors therefore saw different presentations and covered different items.
\end{minipage}\par\vspace{4pt}
\fontsize{9}{10.8}\selectfont
\setlength{\tabcolsep}{3pt}
\setlength{\GTusable}{\dimexpr\textwidth-12\tabcolsep\relax}
\renewcommand{\arraystretch}{1.12}
\begin{tabular}{@{}>{\raggedright\arraybackslash}p{0.290000\GTusable}>{\raggedright\arraybackslash}p{0.105000\GTusable}>{\raggedright\arraybackslash}p{0.121000\GTusable}>{\raggedright\arraybackslash}p{0.121000\GTusable}>{\raggedright\arraybackslash}p{0.121000\GTusable}>{\raggedright\arraybackslash}p{0.121000\GTusable}>{\raggedright\arraybackslash}p{0.121000\GTusable}@{}}
\toprule
\textbf{Question} & \textbf{A--B} & \textbf{A--Astra} & \textbf{A--Qwen} & \textbf{B--Astra} & \textbf{B--Qwen} & \textbf{Astra--Qwen} \\
\midrule
Q2 visible before & 66/\allowbreak{}80 & 59/\allowbreak{}80 & 59/\allowbreak{}80 & 70/\allowbreak{}80 & 67/\allowbreak{}80 & 66/\allowbreak{}80 \\
Q2 visible after & 70/\allowbreak{}80 & 71/\allowbreak{}80 & 72/\allowbreak{}80 & 74/\allowbreak{}80 & 76/\allowbreak{}80 & 76/\allowbreak{}80 \\
Q2 crossing & 56/\allowbreak{}80 & 56/\allowbreak{}80 & 57/\allowbreak{}80 & 54/\allowbreak{}80 & 72/\allowbreak{}80 & 53/\allowbreak{}80 \\
Q3 occurrence & 76/\allowbreak{}92 & 56/\allowbreak{}92 & 68/\allowbreak{}82 & 60/\allowbreak{}92 & 67/\allowbreak{}82 & 47/\allowbreak{}82 \\
Q3 direction & 72/\allowbreak{}92 & 55/\allowbreak{}92 & 44/\allowbreak{}82 & 57/\allowbreak{}92 & 45/\allowbreak{}82 & 25/\allowbreak{}82 \\
Q3 same person & 82/\allowbreak{}92 & 61/\allowbreak{}92 & 66/\allowbreak{}82 & 63/\allowbreak{}92 & 66/\allowbreak{}82 & 49/\allowbreak{}82 \\
Q3 timing determinable & 80/\allowbreak{}92 & 52/\allowbreak{}92 & 72/\allowbreak{}82 & 56/\allowbreak{}92 & 71/\allowbreak{}82 & 45/\allowbreak{}82 \\
\bottomrule
\end{tabular}
\end{table*}

We recomputed pairwise agreement from the original saved answers on the shared items listed, retaining the technically missing responses. These counts describe agreement under the recorded viewing conditions. Human-support classifications use the recorded human responses under the assistance conditions documented in Appendix C. The auxiliary agreement statistics compare the corresponding saved human and/or AI answers. AI visual answers were not used as reference labels or inputs to event matching.

\section*{Appendix E. BASE missed crossings and repair rules}

\label{app:E}

Main-paper Figure 3 connects two BASE misses to their saved track and counting records. In the first case, the track produces a possible crossing for person 110 at the middle horizontal gate of MOT20-02 one frame from the reference, and the track continues. This crossing fails the depth check because distance into the destination region divided by predicted body height is 0.194--0.199, just below 0.2. Dividing the same distance by the output box height would give 0.202--0.208. In the second case, the counting rules find no possible crossing. Although 36 output boxes linked to the person surround the reference, all positions on the 72-frame track lie in the destination region. No transition between regions is observed, and the saved records rule out a gap inside this track. Retaining possible crossings that fail the depth check can affect the first case. The second case has no possible crossing to retain.

\begin{table*}[!tp]
\centering
\small
\begin{minipage}{\textwidth}
The published BASE outputs have uneven recovery across the two development sequences. Both coordinate interpretations are retained below. These per-sequence scores were independently recomputed from the saved outputs.
\end{minipage}\par\vspace{4pt}
\fontsize{9}{10.8}\selectfont
\setlength{\tabcolsep}{3pt}
\setlength{\GTusable}{\dimexpr\textwidth-12\tabcolsep\relax}
\renewcommand{\arraystretch}{1.12}
\begin{tabular}{@{}>{\raggedright\arraybackslash}p{0.180000\GTusable}>{\raggedright\arraybackslash}p{0.170000\GTusable}>{\raggedright\arraybackslash}p{0.120000\GTusable}>{\raggedright\arraybackslash}p{0.140000\GTusable}>{\raggedright\arraybackslash}p{0.140000\GTusable}>{\raggedright\arraybackslash}p{0.120000\GTusable}>{\raggedright\arraybackslash}p{0.130000\GTusable}@{}}
\toprule
\textbf{Coordinates} & \textbf{Sequence} & \textbf{Low} & \textbf{Mid-low} & \textbf{Mid-high} & \textbf{High} & \textbf{Precision} \\
\midrule
One-based & MOT20-01 & 25/\allowbreak{}32 & 14/\allowbreak{}16 & 12/\allowbreak{}13 & 12/\allowbreak{}13 & 63/\allowbreak{}64 \\
One-based & MOT20-02 & 89/\allowbreak{}148 & 141/\allowbreak{}155 & 139/\allowbreak{}150 & 197/\allowbreak{}197 & 566/\allowbreak{}595 \\
Zero-based & MOT20-01 & 24/\allowbreak{}32 & 15/\allowbreak{}16 & 13/\allowbreak{}13 & 12/\allowbreak{}13 & 64/\allowbreak{}65 \\
Zero-based & MOT20-02 & 91/\allowbreak{}148 & 141/\allowbreak{}155 & 139/\allowbreak{}150 & 197/\allowbreak{}197 & 568/\allowbreak{}594 \\
\bottomrule
\end{tabular}
\end{table*}

The combined low-visibility counts of 114--115/180 include these different sequence conditions. As described in main-paper Section 4, BASE uses parameters fitted on MOT20 and processes trajectories offline.

The first row of main-paper Table 3 lists BASE's development misses by stage under both coordinate interpretations. Main-paper Table 4 gives the corresponding repair counts. Appendix K lists the stages for other systems and sequences. The records below explain the remaining cases and how the repair selects possible crossings.

The saved records show no case where a final report met a missed reference's matching conditions but another pairing prevented the match. They also contain no possible crossing that met those conditions and passed the depth check but was removed by R1/R2 in this population. The second category contains misses where boxes linked to the person are present but their tracks yield no possible crossing. A possible crossing assigned to the same person can fail the direction, time or position conditions, including when it is far from the reference time. Some rows in the published trajectories are interpolated. We retain their source information and do not count them as raw detector observations.

The repair looks for a YOLOX-composite event at the same gate and in the same direction, within 1 s and one predicted body height of a BASE possible crossing that failed the depth check. The height comes from the fitted model. Paired boxes must be each other's best match, with overlap at least 0.5 and margins of at least 0.1 over competing boxes in both directions. The rule also requires at least three compatible observations on each side of the crossing time, support on more than half of the frames where both tracks have output, and unique support. Reference person-IDs enter only the evaluation after selection. Because the selection rule was chosen after inspecting development failures, this is an exploratory repair.

The two coordinate versions add 25/21 final events and 14/12 matches, leaving 11/9 added reports unmatched. The low-visibility gain is 6/5, with all old matches retained. Main-paper Section 5.3 identifies added unmatched reports compatible with reference transitions excluded by the depth rule. We keep the precision definition and reference set fixed; additional final reports increase the denominator. Precision falls, so the repair fails the no-precision-loss condition.

Four low-visibility crossings supported after joint review illustrate the remaining misses. Case 1 has no nearby output box that can be linked to the person. Case 2 has the track entirely in the destination region described in main-paper Section 5.2 and Figure 1. Case 3 has ten boxes linked to the person before the reference and none after. Case 4 has final events whose person-ID votes conflict or identify someone else. Cases 1 and 2 also have initial human support. Their recorded timing intervals conflict or are missing.

\section*{Appendix F. Stadium context and reproduction}

\label{app:F}

\subsection*{F.1. Stadium results}

\begin{table*}[!tp]
\centering
\small
\begin{minipage}{\textwidth}
The stadium development study evaluates 54 video time intervals, pairing crossings by time and position. It has no annotations that follow each person's ID across frames. Table F1 reports matched counts, events counted within each interval, and precision. A match may use an event just outside the interval if it falls within the matching tolerance, while the event count excludes that event. Precision is reported directly from the original scorer.
\end{minipage}\par\vspace{4pt}
\fontsize{9}{10.8}\selectfont
\setlength{\tabcolsep}{3pt}
\setlength{\GTusable}{\dimexpr\textwidth-8\tabcolsep\relax}
\renewcommand{\arraystretch}{1.12}
\begin{tabular}{@{}>{\raggedright\arraybackslash}p{0.190000\GTusable}>{\raggedright\arraybackslash}p{0.215000\GTusable}>{\raggedright\arraybackslash}p{0.200000\GTusable}>{\raggedright\arraybackslash}p{0.200000\GTusable}>{\raggedright\arraybackslash}p{0.195000\GTusable}@{}}
\toprule
\textbf{System} & \textbf{Clear: matched; events; precision} & \textbf{Partly obscured} & \textbf{Mostly obscured} & \textbf{All} \\
\midrule
P1 & 108/\allowbreak{}129; 111; .92 & 153/\allowbreak{}201; 164; .91 & 9/\allowbreak{}138; 17; .53 & 270/\allowbreak{}468; 292; .89 \\
CrowdDet + stitching & 104/\allowbreak{}129; 114; .88 & 166/\allowbreak{}201; 193; .84 & 88/\allowbreak{}138; 232; .38 & 358/\allowbreak{}468; 539; .65 \\
CrowdDet composite & 104/\allowbreak{}129; 114; .88 & 171/\allowbreak{}201; 196; .86 & 101/\allowbreak{}138; 296; .34 & 376/\allowbreak{}468; 606; .61 \\
YOLOX composite & 103/\allowbreak{}129; 107; .93 & 165/\allowbreak{}201; 180; .90 & 40/\allowbreak{}138; 87; .45 & 308/\allowbreak{}468; 374; .80 \\
\bottomrule
\end{tabular}
\end{table*}

The stadium study motivated the public-data evaluation that checks which person crossed. Stadium pairing uses time and position, whereas MOT20 matching also checks the person.

\subsection*{F.2. Reproduction instructions and terminology}

The standalone development resource uses Python 3.10.18 and NumPy 2.2.6 in its recorded reference environment. The reconstruction command rebuilds crossing labels and scores the supplied composite example. The tracker command reads compatible MOTChallenge files, derives crossing reports, and matches them to references. Both require official annotations supplied by the user and write into a new output directory. The published-source checkout uses a separate wrapper layout described in its README.

\begin{quote}\small\raggedright

{\GTcode python\allowbreak{} {\normalfont-}B\allowbreak{} reproduce.py\allowbreak{} {\normalfont-}{\normalfont-}gt{\normalfont-}root\allowbreak{} GT\_ROOT\allowbreak{} {\normalfont-}{\normalfont-}seqinfo{\normalfont-}root\allowbreak{} SEQINFO\_ROOT\allowbreak{} {\normalfont-}{\normalfont-}out\allowbreak{} NEW\_OUTPUT}\par

{\GTcode python\allowbreak{} {\normalfont-}B\allowbreak{} score\_tracker.py\allowbreak{} {\normalfont-}{\normalfont-}gt{\normalfont-}root\allowbreak{} GT\_ROOT\allowbreak{} {\normalfont-}{\normalfont-}seqinfo{\normalfont-}root\allowbreak{} SEQINFO\_ROOT\allowbreak{} {\normalfont-}{\normalfont-}tracks\allowbreak{} TRACKS\_DIRECTORY\allowbreak{} {\normalfont-}{\normalfont-}out\allowbreak{} NEW\_OUTPUT}\par

\end{quote}

\begin{table*}[!tp]
\centering
\small
\begin{minipage}{\textwidth}
A fresh clone on the same machine reproduced the composite example at 603/665 precision and the examples for the three trackers sharing detections at low-visibility counts 83/70/84. The generic tracker interface takes event points from output boxes. Reconstructing the original composite counting rules also needs the associated input detections described in main-paper Section 4.
\end{minipage}\par\vspace{4pt}
\fontsize{9}{10.8}\selectfont
\setlength{\tabcolsep}{3pt}
\setlength{\GTusable}{\dimexpr\textwidth-2\tabcolsep\relax}
\renewcommand{\arraystretch}{1.12}
\begin{tabular}{@{}>{\raggedright\arraybackslash}p{0.180000\GTusable}>{\raggedright\arraybackslash}p{0.820000\GTusable}@{}}
\toprule
\textbf{Label} & \textbf{Meaning} \\
\midrule
E1 & Automatic assignment of failure categories in a fixed order from saved records \\
E2 & Deleting equal numbers of observations at dispersed or consecutive positions \\
E3 & Range of bin counts across equally optimal event matches \\
E4 & Standard tracking metrics alongside event metrics \\
H-A & Low/\allowbreak{}high recall inequalities fixed before each confirmation sequence is evaluated \\
H-B & Composite low-visibility count exceeds each of the three trackers sharing detections on each confirmation sequence \\
Q2 & Human judgments of visibility and crossing on development events selected by system outcomes \\
Q3 & Human judgments of reference support on development items sampled by sequence and visibility \\
M3 /\allowbreak{} M4 & Admitting low-score detections /\allowbreak{} stitching track fragments \\
C1 /\allowbreak{} R1 /\allowbreak{} R2 & Checking depth /\allowbreak{} grouping events from the same track /\allowbreak{} checking which regions the observations occupy before and after \\
\bottomrule
\end{tabular}
\end{table*}

\subsection*{F.3. Detailed geometry and numerical conventions}

For each sequence, ordinary least squares fits body height h from box-center x-coordinate x\_c and top y-coordinate y\textsubscript{0}:

{\GTcode h = a + b x\_c/W + c y₀/H;     ĥ(x,y) = (a + b x/W + c y/H)/(1 + c/H).}

The second expression predicts height at a bottom point (x,y). We fit it using evaluated pedestrian boxes strictly inside all four image boundaries, without a visibility threshold. Horizontal gates lie at 0.30H, 0.50H, and 0.70H, spanning 0.02W to 0.98W. Vertical gates lie at 0.25W, 0.50W, and 0.75W, spanning 0.02H to 0.98H. At midpoint height h\_m, the adjacent rectangular regions have gap half-width 0.05h\_m and depth h\_m. IN denotes movement from the upper or left region to the other; OUT denotes the reverse.

A valid point entering the region opposite its last assigned region creates a transition. Points outside both rectangles leave the remembered region unchanged. An absence of track observations longer than 0.5 s resets the fragment. The entry frame and position come from an observed point, without interpolation. To decide whether a reference meets the depth rule, we search the person's available rows from entry through 50 frames later, inclusive. A point in the destination region meets that rule when its Euclidean distance from the finite gate-facing edge reaches 0.2 times that row's annotated height. This search is not restricted to the same fragment. The timestamp is the observed entry into the opposite region across the gap around the central line.

Height fitting requires at least three eligible rows, matrix rank three, and finite coefficients. Inverting the geometry requires a finite nonzero denominator and a positive midpoint height. For system C1, a nonpositive or nonfinite denominator or height makes normalized depth unavailable. Extrapolated values are flagged and retained without clipping. The reference depth rule adds a pixel tolerance of 10\textsuperscript{-}\textsuperscript{9} to the edge distance before comparing it with 0.2 times annotated height; the system depth check applies its tolerance to normalized depth. We attribute no numerical effect to this tolerance difference. An invalid event point leaves the remembered region unchanged but updates the observation time. Event points exclude side and lower boundaries; height fitting also excludes the upper boundary.

\subsection*{F.4. Artifact capabilities and availability}

\begin{table*}[!tp]
\centering
\small
\fontsize{9}{10.8}\selectfont
\setlength{\tabcolsep}{3pt}
\setlength{\GTusable}{\dimexpr\textwidth-4\tabcolsep\relax}
\renewcommand{\arraystretch}{1.12}
\begin{tabular}{@{}>{\raggedright\arraybackslash}p{0.220000\GTusable}>{\raggedright\arraybackslash}p{0.320000\GTusable}>{\raggedright\arraybackslash}p{0.460000\GTusable}@{}}
\toprule
\textbf{Evidence layer} & \textbf{Implemented capability} & \textbf{Scope and required material} \\
\midrule
Standalone resource & Rebuild labels,\allowbreak{} derive crossing reports from output tracks,\allowbreak{} match reports to the correct person and compute ranges across equally optimal matches & MOT20-01/\allowbreak{}02; user-supplied official GT and sequence metadata; compatible tracks or supplied examples \\
Generic inspector & Final matches,\allowbreak{} nearby events,\allowbreak{} reference sources,\allowbreak{} and nearby output boxes linked to the person in space and time & Saved matching and checks linking boxes to people in each frame plus source tracks; human support is supplied separately \\
Counting-rule records & Possible crossings before filtering,\allowbreak{} C1 check records,\allowbreak{} groups and final events & Saved by the scorer; checking the person-ID of a possible crossing that was rejected requires an additional study-specific link to person-ID records \\
BASE diagnosis and repair study & Complete list of missed crossings by stage,\allowbreak{} person-ID checks for possible crossings,\allowbreak{} and counts of added reports & Separate study scripts and saved numerical records; stage and repair counts independently recomputed; links to human judgments and later calculations checked through additional implementations \\
Human-support study & Summaries of human support for each saved version,\allowbreak{} linked to the reported matches & Separate aggregate research records; individual answers remain private; counts and links reproduced from versioned records \\
\bottomrule
\end{tabular}
\end{table*}

The separate study records are needed to reproduce the study-specific links between outputs, references and human judgments.

Archived study records contain summaries of the misses with source-file hashes, saved C1 records for possible crossings, and the complete list of events added by the repair. They also include original report excerpts for the two initially supported cases. A written report documents the track entirely within the destination region. Reproducing the full BASE workflow requires the source scripts and original inputs identified in these records. A detailed view lists the saved possible crossings eligible for examining low-visibility misses (26 rows), with a source and hash for each. It shows whether the track continues, the person-ID evidence on each side of the crossing time, and paired height ratios where available. Missing fields are marked.

\subsection*{F.5. Related-work scope}

\begin{table*}[tp]
\centering
\small
\fontsize{9}{10.8}\selectfont
\setlength{\tabcolsep}{3pt}
\setlength{\GTusable}{\dimexpr\textwidth-4\tabcolsep\relax}
\renewcommand{\arraystretch}{1.12}
\begin{tabular}{@{}>{\raggedright\arraybackslash}p{0.310000\GTusable}>{\raggedright\arraybackslash}p{0.350000\GTusable}>{\raggedright\arraybackslash}p{0.340000\GTusable}@{}}
\toprule
\textbf{Related line of work} & \textbf{Established measurement or analysis} & \textbf{Focus of the GateTrace study} \\
\midrule
Event-based tracking evaluation [9,\allowbreak{}23] & Matched events and changes in object IDs & Visibility groups and a reproducible record of where each person's crossing is lost \\
Line crossing and pedestrian-flow evaluation [10,\allowbreak{}1,\allowbreak{}15,\allowbreak{}16] & Crossing/\allowbreak{}count accuracy,\allowbreak{} counting geometry,\allowbreak{} and pedestrian-flow measurements & Event records grouped by visibility and misses among human-supported crossings \\
Video individual counting [24,\allowbreak{}25] & Number of distinct pedestrians,\allowbreak{} estimated from inflow and outflow without tracking & Each person's crossing reports derived from tracker output \\
Tracking evaluation [7,\allowbreak{}8,\allowbreak{}2,\allowbreak{}14] & Tracking quality and difficulty & Crossing recovery at specified counting gates under fixed counting rules \\
Error analysis and reference interventions [17,\allowbreak{}18] & Error categories and controlled reference conditions & Losses in counting rules,\allowbreak{} repairs to selected possible crossings,\allowbreak{} and precision cost across all reports \\
\bottomrule
\end{tabular}
\end{table*}

\section*{Appendix G. Event-definition and repair-selection checks}

\label{app:G}

\textbf{The reference rule affects which reports match.} Some unmatched reports may correspond to annotated transitions excluded by the reference depth rule. We keep the reports from the relaxed counting rules fixed and add these excluded transitions to the reference list. This yields 147--151 additional matches for BASE and 121 for ByteTrack. Low-visibility recovery becomes 166--169/239 for BASE and 124/239 for ByteTrack, compared with 243/244 and 239/244 at high visibility; precision is 93.5--94.4\% and 78.9\%, respectively. Separately, 17 of BASE's original final reports that were unmatched under the original definition match newly admitted references. These gains come from scoring existing reports against a larger reference list. Establishing whether the added references represent physical passages requires separate evidence. The precision loss against the original references remains.

The fixed-track comparison uses original definition O, with references retained by the depth rule and final events, and symmetric definition S, with transitions before filtering on both sides. S has 901 references: 90 on 01 and 811 on 02, including 239 low-visibility and 244 high-visibility references. We keep the geometry, bottom-point rules, invalid boundaries and 0.5-second gap rule. Both the system's possible crossings and the reference crossings included in the comparison change; the person, time and distance matching conditions stay fixed.

Added low-visibility reference matches that persist under both BASE coordinate interpretations number zero on 01 and twelve on 02; ByteTrack adds three and nine. Each of the four system--sequence comparisons had to give at least max(2,ceil(0.05 N\_low)) stable additions involving at least two people: two additions on 01 and eight on 02. Three comparisons pass, while BASE/01 fails. We therefore stop before expanding human review. This threshold decides whether to spend resources on further review; it is separate from a statistical significance test.

Scoring the original-reference subset separately and extracting it from full S matching give the same reference memberships in this run. On that subset, BASE matches 127--128/180 low-visibility references and ByteTrack 95/180. The corresponding high-visibility counts are BASE 210/210 and ByteTrack 208/210. Thus the visibility gap remains after removing C1/R1/R2 while keeping the original geometry, person-ID checks, references, and tolerances. Independent recomputation agreed on these high-visibility counts. Across the full S reference set, BASE matches 166--169/239 and ByteTrack 124/239. Main-paper Section 5.3 and Appendix J.5 report precision ranges across the two coordinate interpretations. Seventeen previously unmatched BASE final reports in each interpretation and 21 ByteTrack reports match newly included references. These reports were already present; they differ from the 11/9 unmatched reports added by the earlier BASE repair.

A separate check compares four repair options using existing caches from all four previously analyzed sequences. The baseline A0 for this check uses index-corrected ByteTrack at high/low/new thresholds 0.6/0.1/0.7, no external M3 or M4, and event points from the original detection boxes associated with each track. A1 changes admission and tracking thresholds to the low-score settings without M4. A2 adds M4 to A0. A3 changes A0's event points to the bottom centers of tracker output boxes. Geometry, event rules and scoring stay fixed. This retrospective comparison has its own baseline and leaves the main-paper system rows unchanged.

We accept an action's positive low-visibility gain only if the high-visibility count and precision do not decrease. Precision is computed with matching that requires the correct person. Otherwise we retain A0 and assign zero utility. O chooses the highest accepted utility for each sequence, then averages the four sequence rates equally. C* recommends the same action for every sequence, subject to the same acceptance rule and fallback. Fixed recommendation A1 achieves the same accepted utility as selecting separately by sequence: it is adopted on 01/03/05, with fallback to A0 on 02. Thus O\ensuremath{-}C*=0, below the required one-percentage-point advantage for selection. We stop this check without a human selection experiment. A separate implementation reproduced the event person-ID votes, allowed matching pairs, and acceptance and fallback arithmetic from the saved upstream records, using the same assignment solver.

The definition comparison was independently recomputed. An additional implementation reproduced the repair-selection analysis from saved records. Appendix I describes both checks.

\section*{Appendix H. Details of the counting rules}

\label{app:H}

R1 groups events that passed C1 and have the same track and direction. Successive events may be at most two seconds apart, so a group can span more than two seconds. R2 checks the intervals [earliest\ensuremath{-}1,earliest) and (latest,latest+1] in seconds. It counts only valid observations assigned to either region. If either interval has fewer than three such points, it keeps the group and flags insufficient evidence without checking the proportions in the required regions. Otherwise both required proportions must reach 0.5.

For person-ID voting, we use frames strictly before the system event on the earlier side and frames at or after it on the later side. Both are restricted to the complete \ensuremath{\pm}25-frame window. A person-ID can be assigned from the combined window even when one side has insufficient evidence. If both sides receive different person-IDs, their disagreement takes precedence over the combined assignment. The proportions of observations in each region and the person-ID assigned on each side are recorded separately.

The three height values serve different steps. The reference depth rule uses the annotated height of each supporting box. System C1 uses the height predicted by the fitted model. Final matching uses the annotated height at the reference event frame. Invalid image points, missing information for the person-ID check, and uncertainty are retained at the stages where they occur.

\section*{Appendix I. Reproducing calculations from saved records}

\label{app:I}

The checks distinguish saved inputs, the calculations made from them, and whether a passage physically occurred. Claude (Anthropic) generated and ran a separate implementation (see the LLM Usage Statement) to recompute the BASE scores, repair counts, missed-crossing stages, and symmetric-definition results. Its results matched on 8 November 2026. These implementations share the exact matcher and some person-ID logic with the original pipeline. They reuse the original-definition final events, decisions about whether the depth requirement was met, and some reference attributes from saved records. The checkers could see the target results. Earlier recomputation records cover results for the three trackers sharing saved detections and the confirmation results. On 9 November 2026 the same assistant recomputed Appendix K with a separate implementation of output parsing, assigning person-IDs to boxes and events, matching and stage assignment, while reusing saved counting-rule records. Every bin count, all 7,896 per-reference match states for BASE on 03/05 and all 4,321 per-reference stages behind Table K2 (all visibility bins) agree with the original outputs.

A separate code audit generated and run with OpenAI Codex (see the LLM Usage Statement) rebuilt human-support classifications, time-interval transformations, versioned agreement counts, and their links to full-sequence matches from saved answers and mappings. A separate aggregation rebuilt every human-AI agreement fraction from the individual saved responses. Another implementation rebuilt event person-IDs, allowed matching pairs, all-bin scores, and adoption and fallback calculations for the emitted-box variant and the repair options. It reused the exact assignment solver, saved event extraction, and assignments linking output boxes in each frame to reference people. Figure values and geometry were checked against the corresponding reproduced records. Input hashes, per-item comparisons, and execution records are archived with these checks.

The BASE repair is an exploratory engineering study designed after inspecting development failures. Its acceptance rule was recorded with a hash when execution started, before scores were computed; the protocol also calls the study exploratory. The precision decrease causes it to fail the no-precision-loss condition.

One archived one-based BASE repair run is marked as failed because a console print error occurred after its outputs had been written; these outputs reproduce the reported repair row. The coordinate origin of the published BASE tracks remains unresolved, so both coordinate interpretations are reported.

These checks reproduce the reported calculations from saved inputs. Independent validation of the human judgments and a fully independent end-to-end replication of event extraction and matching remain outstanding.

\section*{Appendix J. Material condensed from the main paper}

\label{app:J}

The following passages give the details condensed out of the main paper, so that every condensed result remains documented.

\subsection*{J.1. Full development table (including P1 and CrowdDet systems)}

\begin{table*}[!tp]
\centering
\small
\begin{minipage}{\textwidth}
\textbf{Table J1. Development results on MOT20-01/02.} Recall cells give matched/total references within each fixed visibility bin. Precision includes all final reports. The listed results have independent numerical recomputations. P1 is the earlier YOLO pipeline; O1 uses reference boxes.
\end{minipage}\par\vspace{4pt}
\fontsize{9}{10.8}\selectfont
\setlength{\tabcolsep}{3pt}
\setlength{\GTusable}{\dimexpr\textwidth-10\tabcolsep\relax}
\renewcommand{\arraystretch}{1.12}
\begin{tabular}{@{}>{\raggedright\arraybackslash}p{0.330000\GTusable}>{\raggedright\arraybackslash}p{0.130000\GTusable}>{\raggedright\arraybackslash}p{0.140000\GTusable}>{\raggedright\arraybackslash}p{0.130000\GTusable}>{\raggedright\arraybackslash}p{0.130000\GTusable}>{\raggedright\arraybackslash}p{0.140000\GTusable}@{}}
\toprule
\textbf{System} & \textbf{Low} & \textbf{Mid-low} & \textbf{Mid-high} & \textbf{High} & \textbf{Precision} \\
\midrule
P1 & 27/\allowbreak{}180 & 76/\allowbreak{}171 & 126/\allowbreak{}163 & 197/\allowbreak{}210 & 426/\allowbreak{}590 \\
CrowdDet /\allowbreak{} P2 & 81/\allowbreak{}180 & 140/\allowbreak{}171 & 144/\allowbreak{}163 & 201/\allowbreak{}210 & 566/\allowbreak{}697 \\
CrowdDet composite & 86/\allowbreak{}180 & 136/\allowbreak{}171 & 144/\allowbreak{}163 & 203/\allowbreak{}210 & 569/\allowbreak{}712 \\
YOLOX + ByteTrack & 83/\allowbreak{}180 & 142/\allowbreak{}171 & 149/\allowbreak{}163 & 206/\allowbreak{}210 & 580/\allowbreak{}644 \\
YOLOX + OC-SORT & 70/\allowbreak{}180 & 137/\allowbreak{}171 & 147/\allowbreak{}163 & 206/\allowbreak{}210 & 560/\allowbreak{}593 \\
YOLOX + BoT-SORT & 84/\allowbreak{}180 & 142/\allowbreak{}171 & 150/\allowbreak{}163 & 208/\allowbreak{}210 & 584/\allowbreak{}634 \\
YOLOX + BoT-SORT + ReID & 81/\allowbreak{}180 & 141/\allowbreak{}171 & 149/\allowbreak{}163 & 208/\allowbreak{}210 & 579/\allowbreak{}628 \\
YOLOX composite & 99/\allowbreak{}180 & 149/\allowbreak{}171 & 150/\allowbreak{}163 & 205/\allowbreak{}210 & 603/\allowbreak{}665 \\
GT-body input /\allowbreak{} O1 & 173/\allowbreak{}180 & 168/\allowbreak{}171 & 161/\allowbreak{}163 & 209/\allowbreak{}210 & 711/\allowbreak{}723 \\
\bottomrule
\end{tabular}
\end{table*}

\subsection*{J.2. System and confirmation details}

The three trackers share the same saved detections from the official {\GTcode bytetrack\_x\_mot17} YOLOX-X checkpoint distributed with ByteTrack [3,19]. They are ByteTrack, OC-SORT, and BoT-SORT without ReID [3,4,5]. The documented detector training excludes MOT20. Export uses 800\ensuremath{\times}1440 inputs, score threshold 0.01, and NMS 0.7. Each tracker uses its authors' fixed MOT17 settings, including its own admission threshold, output filtering, and camera-motion handling.

The YOLOX composite admits detections at 0.10, runs index-corrected ByteTrack at high/low/new thresholds 0.25/0.10/0.25, then stitches fragments (M4) and applies the counting rules. The CrowdDet composite applies the same later steps to CrowdDet inputs [20]. P2 is the original CrowdDet baseline with admission at 0.25. O1 feeds reference full-body boxes to the fixed tracker and counting rules. The study trains no model weights.

The fixed 03/05 confirmation includes seven systems: the three trackers sharing detections, the two composites, P2, and O1. Four systems share one archived YOLOX export per sequence. We recorded hashes of inputs, code, parameters, environment, and outputs for the execution. Results depend on these cached detections because floating-point inference can vary. Earlier P1/P2 results and O1/O2 diagnostics on 03/05 had been inspected. The later seven-system execution followed a protocol fixed beforehand.

For each sequence $s$, H-A requires each tracker $m$ among the three sharing detections to satisfy $10K_{low,s}(m)\leq7N_{low,s}$ and $10K_{high,s}(m)\geq9N_{high,s}$. H-B requires the YOLOX composite's low-visibility count to exceed the largest count among those trackers. Both sequences must pass; their combined values are descriptive. The BASE development results, repair and definition comparisons are development analyses. BASE's 03/05 scores are an additional analysis (Appendix K).

BASE contributes published trajectories [6]. Its setup includes a separate detector recipe, MOT20-based parameter estimation and tuning, look-ahead, and interpolation. We retain both one-based and zero-based coordinate interpretations. The release identifies the files' source but does not link each file to a specific leaderboard submission.

An exploratory repair retains possible crossings from BASE that failed the depth check when YOLOX-composite output meets the support rules. It preserves each possible crossing's person-ID, time, and position. Selection uses output geometry and mutual track compatibility; reference person-IDs enter only the later evaluation. We designed the selection rule after inspecting development failures. Before this exploratory run, we recorded an acceptance rule requiring at least three additional low-visibility matches under both coordinate interpretations, no high-visibility loss in either sequence, and no overall precision loss.

A separate fixed-track comparison uses original definition O and symmetric region transitions S. O pairs final system events with references retained by the depth rule. S pairs possible crossings from system outputs before filtering with all reference transitions before the depth rule, keeping repeated traversals on both sides and removing depth checks and R1/R2. The person-ID checks and matching tolerances stay fixed. We report the original-reference subset separately from the full alternative reference set. A threshold fixed before the check determines whether to proceed to additional human review.

\subsection*{J.3. Human-review design}

Q2 samples 40 low-visibility events missed by five earlier systems and 40 recovered by at least three. BASE entered the study later. The two people who reviewed the videos see the target and video, with system outputs and group labels hidden. They judge whether any body part is visible before and after, and whether the video supports a crossing.

Q3 samples by sequence and visibility after excluding Q2. Its eligible population has 644 references retained by the depth rule, including 100 low-visibility events. Its 93 items include 88 references from that list, four transitions excluded by that rule, and one open-search window. The sample from the reference list has 34 low-visibility events and 18 in each other bin. Before responses, we randomly added sixteen low-visibility MOT20-02 items under a recorded rule. The two people who reviewed the videos judge whether a crossing occurred, its direction, whether the person remained the same, and whether its time can be determined. They may record supporting time intervals.

The two people who reviewed the videos submitted initial answers separately, then discussed disagreements. Person A discussed definitions and illustrated examples with an assistant while completing the initial answers; the joint-review page displayed AI answers for reference. The primary analysis uses initial answers. Q3 support requires both people to endorse occurrence, agree that the person stayed the same, and give a direction compatible with the reference. An explicit rejection of occurrence or of the person remaining the same, or an incompatible direction, counts as a conflict. Incomplete or uncertain support remains unresolved. We link these judgments to saved matches at the fixed reference times.

\subsection*{J.4. Repair accounting}

The repair adds six or five low-visibility matches, keeps all earlier matches, and leaves high-visibility counts unchanged. Across all bins, it adds 25 or 21 final reports and 14 or 12 matched references. Eleven or nine additions remain unmatched. Of these, 9/11 and 8/9 are compatible with saved transitions excluded from the reference list by the depth rule. Each coordinate interpretation also adds one report under the category that ignores person-IDs; the one-based version adds one report without an eligible same-person reference transition.

Precision falls under both coordinate interpretations, so the repair fails the protocol's no-precision-loss condition. Most added unmatched reports align with transitions excluded by the reference depth rule. This explains their precision cost under the fixed reference list. Whether those reports describe useful physical passages needs further evidence. Main-paper Figure 3 shows the two kinds of remaining miss: a possible crossing that fails the depth check and a track in which the counting rules find no possible crossing.

\subsection*{J.5. Per-sequence effect of the relaxed definition}

Recall remains lower at low visibility under S. Precision changes from 95.45--95.90\% to 93.55--94.44\% for BASE and from 90.06\% to 78.86\% for ByteTrack. Seventeen existing unmatched BASE final reports under each coordinate interpretation, and 21 ByteTrack reports, match newly included references. Both retaining possible crossings and expanding the reference list therefore contribute to the changed scores.

The effect varies by sequence. Added low-visibility matches that persist under both BASE coordinate interpretations number zero on 01 and twelve on 02; ByteTrack adds three and nine. The threshold for further human review required at least two on 01 and eight on 02 for each system, involving at least two people in each comparison. Three of four comparisons passed, so the study stopped before expanding human review. The scores are sensitive to the definition under the tested outputs, but the required improvement across all conditions was not met.

\subsection*{J.6. Outcome-selected visibility audit (Q2)}

In Q2, both people initially support the crossing for 18/40 items missed by all earlier systems and 35/40 controls. For each supported item, both report some body part visible before and after. The prespecified complete-invisibility criterion fails: the common-miss subset has fewer than the required twenty supported passages, and the invisibility bounds fail their conditions. The supported subset therefore contains visible information. Unsupported and uncertain items remain in the original sample. Seeing a body part does not establish the precise foot position or event time.

\subsection*{J.7. Diagnostic controls, matching ambiguity, and frame metrics}

Deleting observations changes recovery even when their number is held fixed. At requested retention 0.5, dispersed deletion gives overall recall near 0.896 on both development sequences; contiguous deletion gives approximately 0.224--0.225. Deleting thirteen consecutive observations leaves 0.56 s between the retained endpoints, exceeding the 0.5 s reset in the counting rules. The curves therefore reflect where observations are missing in time and how the complete processing chain responds.

The head-observation diagnostic, check of possible crossings across gaps, and relinking test using reference person-IDs all fail their recorded conditions for proceeding. Fixed BoT-SORT ReID changes low-visibility recovery from 84 to 81. Removing M4 from the YOLOX composite gives 97/180, against 99/180 for the full configuration. Independent checks cover these results within the scope described in Appendix B.2. The head diagnostic uses the same-video head annotations described by Sundararaman et al. [22].

All equally optimal matches give the same bin counts in the primary mode that checks person-IDs; counts can differ when person-IDs are ignored. Standard frame metrics also decrease at low visibility. Across the two confirmation sequences, three of 56 metric comparisons meet the recorded thresholds for a ranking reversal. All concern ByteTrack and BoT-SORT on sequence 05. BoT-SORT scores higher on HOTA, IDF1, and low-visibility frame person-ID recall; ByteTrack recovers five more low-visibility crossings, exactly the required difference. The sequence-03 comparisons do not meet these thresholds. This is a local example of the metrics ranking systems differently, with limited evidence for a broader ranking effect.

\subsection*{J.8. Artifact scope and data handling}

The released resource lets readers inspect the reported results. Its standalone development tools rebuild 01/02 labels and score compatible output tracks. The inspector shows final matches and nearby output boxes linked to the person. Confirmation outputs, counting-rule records, and later analyses linking BASE and human judgments are archived separately. Fresh-clone checks ran on the same machine; reproduction on a different machine has not been tested. The four-option repair check found no advantage over the best fixed recommendation under the same per-sequence acceptance rule, and stopped (Appendix G). The recorded stages identify which operation could affect a case. The repair failed its adoption criterion and the selection check found no advantage. The value of these records for choosing repairs remains unresolved.

The public-data study uses MOT20 under its documented terms. Derived labels retain CC BY-NC-SA 3.0 attribution; code and upstream components retain their respective licenses. The research package contains numerical derivatives and reconstruction tools. Users obtain original public videos and model weights from their providers. Private stadium videos and individual human answers remain outside the public package. Person-IDs are dataset track IDs. Stadium results pair crossings by time and position because person-ID ground truth is unavailable.

\section*{Appendix K. Analyses added after the fixed confirmation}

\label{app:K}

We ran two additional analyses on 9 November 2026, after the fixed confirmation in main-paper Section 4 and the development analyses. Both preserve the confirmation decision and all earlier scores.

\subsection*{K.1. BASE on MOT20-03/05}

The BASE release used for the development trajectories [6] also provides MOT20-03 and MOT20-05 trajectories at the same repository commit. We scored them against the confirmation references with the counting rules, person-ID checks and matching unchanged. Before scoring 03/05, the new input adapter had to reproduce BASE's saved development results under both coordinate interpretations. Every gate's events, person-IDs assigned to boxes in each frame, matches and scores agreed. BASE's parameters were fitted on the MOT20 training sequences, including 03 and 05. Table K1 therefore reports outputs on sequences already used by BASE. These are additional results, outside the held-out setting and outside H-A and H-B.

On MOT20-03/05, BASE recovers a larger fraction of low-visibility crossings than on the development sequences (78.2--78.5\% against 63.3--63.9\%), and a gap of about 21 points to high visibility remains.

\begin{table*}[!tp]
\centering
\small
\begin{minipage}{\textwidth}
\textbf{Table K1. BASE on MOT20-03/05 (additional analysis).} Matched/total references per visibility bin; precision counts all final reports.
\end{minipage}\par\vspace{4pt}
\fontsize{9}{10.8}\selectfont
\setlength{\tabcolsep}{3pt}
\setlength{\GTusable}{\dimexpr\textwidth-12\tabcolsep\relax}
\renewcommand{\arraystretch}{1.12}
\begin{tabular}{@{}>{\raggedright\arraybackslash}p{0.160000\GTusable}>{\raggedright\arraybackslash}p{0.100000\GTusable}>{\raggedright\arraybackslash}p{0.140000\GTusable}>{\raggedright\arraybackslash}p{0.150000\GTusable}>{\raggedright\arraybackslash}p{0.150000\GTusable}>{\raggedright\arraybackslash}p{0.150000\GTusable}>{\raggedright\arraybackslash}p{0.150000\GTusable}@{}}
\toprule
\textbf{Reading} & \textbf{Sequence} & \textbf{Low} & \textbf{Mid-low} & \textbf{Mid-high} & \textbf{High} & \textbf{Precision} \\
\midrule
One-based & 03 & 104/\allowbreak{}136 & 329/\allowbreak{}339 & 384/\allowbreak{}391 & 568/\allowbreak{}572 & 1385/\allowbreak{}1430 \\
One-based & 05 & 231/\allowbreak{}291 & 784/\allowbreak{}804 & 748/\allowbreak{}770 & 640/\allowbreak{}645 & 2403/\allowbreak{}2467 \\
One-based & 03+05 & 335/\allowbreak{}427 & 1113/\allowbreak{}1143 & 1132/\allowbreak{}1161 & 1208/\allowbreak{}1217 & 3788/\allowbreak{}3897 \\
Zero-based & 03 & 104/\allowbreak{}136 & 330/\allowbreak{}339 & 384/\allowbreak{}391 & 567/\allowbreak{}572 & 1385/\allowbreak{}1429 \\
Zero-based & 05 & 230/\allowbreak{}291 & 782/\allowbreak{}804 & 748/\allowbreak{}770 & 637/\allowbreak{}645 & 2397/\allowbreak{}2462 \\
Zero-based & 03+05 & 334/\allowbreak{}427 & 1112/\allowbreak{}1143 & 1132/\allowbreak{}1161 & 1204/\allowbreak{}1217 & 3782/\allowbreak{}3891 \\
\bottomrule
\end{tabular}
\end{table*}

By sequence, BASE's low-visibility recall is 75.0--78.1\% on 01, 60.1--61.5\% on 02, 76.5\% on 03 and 79.0--79.4\% on 05 (01 and 02 from Appendix E). MOT20-02 contains 148 of the 180 low-visibility development references and has the lowest recall.

\subsection*{K.2. Stage breakdown for further systems}

We followed the order in main-paper Section 3.4 when checking saved records from the three trackers sharing detections on all four sequences, the YOLOX composite on 03/05, and BASE on 03/05. The categories, their order, the \ensuremath{\pm}25-frame window and matching tolerances remain the same as for BASE on the development sequences; Table K2 repeats those development counts. Records for the three trackers on 01/02 come from the run used in main-paper Table 1, and 03/05 records come from the confirmation run. In every comparison, the saved low-visibility match count agrees with main-paper Table 1, main-paper Table 2 or Table K1. We omit the composite on 01/02 because its saved counting-rule records come from a later regression run with one mid-low match different from main-paper Table 1.

\begin{table*}[!tp]
\centering
\small
\begin{minipage}{\textwidth}
\textbf{Table K2. Where low-visibility crossings were missed, by system and sequence.} A: no nearby output box can be linked to the person. B: nearby boxes are linked to the person, but the counting rules find no possible crossing. C: a possible crossing meets the matching conditions but fails the depth check. D: a possible crossing is not accepted as belonging to the person in the reference. E: a possible crossing assigned to the same person fails the direction, time or position conditions. F: a possible crossing passed the depth check but was removed when merging events or checking positions before and after them. G: another mismatch involving a possible crossing from a track containing nearby boxes linked to the person. No missed crossing has a final report that meets the matching conditions. Main-paper Table 3 combines sequences and merges F and G.
\end{minipage}\par\vspace{4pt}
\fontsize{9}{10.8}\selectfont
\setlength{\tabcolsep}{3pt}
\setlength{\GTusable}{\dimexpr\textwidth-18\tabcolsep\relax}
\renewcommand{\arraystretch}{1.12}
\begin{tabular}{@{}>{\raggedright\arraybackslash}p{0.244898\GTusable}>{\raggedright\arraybackslash}p{0.081633\GTusable}>{\raggedright\arraybackslash}p{0.081633\GTusable}>{\raggedright\arraybackslash}p{0.081633\GTusable}>{\raggedright\arraybackslash}p{0.081633\GTusable}>{\raggedright\arraybackslash}p{0.081633\GTusable}>{\raggedright\arraybackslash}p{0.081633\GTusable}>{\raggedright\arraybackslash}p{0.081633\GTusable}>{\raggedright\arraybackslash}p{0.081633\GTusable}>{\raggedright\arraybackslash}p{0.102041\GTusable}@{}}
\toprule
\textbf{System} & \textbf{Sequence} & \textbf{A} & \textbf{B} & \textbf{C} & \textbf{D} & \textbf{E} & \textbf{F} & \textbf{G} & \textbf{Total} \\
\midrule
BASE,\allowbreak{} one-based & 01 & 3 & 4 & 0 & 0 & 0 & 0 & 0 & 7 \\
BASE,\allowbreak{} one-based & 02 & 17 & 19 & 13 & 5 & 5 & 0 & 0 & 59 \\
BASE,\allowbreak{} one-based & 03 & 11 & 7 & 11 & 1 & 2 & 0 & 0 & 32 \\
BASE,\allowbreak{} one-based & 05 & 15 & 21 & 14 & 7 & 3 & 0 & 0 & 60 \\
BASE,\allowbreak{} zero-based & 01 & 3 & 4 & 1 & 0 & 0 & 0 & 0 & 8 \\
BASE,\allowbreak{} zero-based & 02 & 17 & 19 & 12 & 4 & 5 & 0 & 0 & 57 \\
BASE,\allowbreak{} zero-based & 03 & 10 & 8 & 10 & 2 & 2 & 0 & 0 & 32 \\
BASE,\allowbreak{} zero-based & 05 & 15 & 21 & 15 & 8 & 2 & 0 & 0 & 61 \\
ByteTrack & 01 & 3 & 7 & 3 & 1 & 0 & 0 & 1 & 15 \\
ByteTrack & 02 & 16 & 23 & 8 & 19 & 9 & 1 & 6 & 82 \\
ByteTrack & 03 & 27 & 23 & 9 & 12 & 5 & 0 & 4 & 80 \\
ByteTrack & 05 & 90 & 29 & 25 & 42 & 7 & 1 & 13 & 207 \\
OC-SORT & 01 & 5 & 7 & 2 & 2 & 0 & 0 & 1 & 17 \\
OC-SORT & 02 & 19 & 35 & 14 & 7 & 10 & 0 & 8 & 93 \\
OC-SORT & 03 & 46 & 30 & 14 & 5 & 4 & 0 & 0 & 99 \\
OC-SORT & 05 & 119 & 81 & 23 & 12 & 4 & 0 & 10 & 249 \\
BoT-SORT & 01 & 3 & 6 & 4 & 3 & 0 & 1 & 0 & 17 \\
BoT-SORT & 02 & 15 & 21 & 8 & 18 & 9 & 0 & 8 & 79 \\
BoT-SORT & 03 & 30 & 24 & 10 & 9 & 4 & 0 & 2 & 79 \\
BoT-SORT & 05 & 88 & 40 & 28 & 39 & 9 & 0 & 8 & 212 \\
YOLOX composite & 03 & 17 & 22 & 12 & 13 & 5 & 0 & 0 & 69 \\
YOLOX composite & 05 & 37 & 56 & 30 & 28 & 11 & 1 & 11 & 174 \\
\bottomrule
\end{tabular}
\end{table*}

Across 03/05, the composite misses 243 low-visibility crossings. Categories A and B account for 132 (54.3\%), and category C for 42 (17.3\%). By sequence, A and B account for between 45.6\% of misses (BoT-SORT on 02) and all seven (BASE on 01, one-based reading). Category C accounts for at most 34.4\% (BASE on 03, one-based reading).

\end{document}